\documentclass[lettersize,journal]{IEEEtran}
\usepackage{amsmath,amsfonts}
\usepackage{algorithmic}
\usepackage{algorithm}
\usepackage{array}
\usepackage[caption=false,font=normalsize,labelfont=sf,textfont=sf]{subfig}
\usepackage{textcomp}
\usepackage{stfloats}
\usepackage{url}
\usepackage{verbatim}
\usepackage{graphicx}
\usepackage{cite}
\begin{document}

\title{DFCANet: Dense Feature Calibration-Attention Guided Network for Cross Domain Iris Presentation Attack Detection}

\author{
    \IEEEauthorblockN{Gaurav Jaswal\IEEEauthorrefmark{1}, Aman Verma\IEEEauthorrefmark{2}, Sumantra Dutta Roy\IEEEauthorrefmark{1}, Raghavendra Ramachandra\IEEEauthorrefmark{3}}\\
    \IEEEauthorblockA{\IEEEauthorrefmark{1}Indian Institute of Technology Delhi, India}\\
       \IEEEauthorblockA{\IEEEauthorrefmark{2}National Institute of Technology Raipur, India}\\
    \IEEEauthorblockA{\IEEEauthorrefmark{3}Norwegian University of Science and Technology Norway, Norway\\
    Email: \IEEEauthorrefmark{1}(gauravjaswal; sumantra)@iitd.ac.in,
\IEEEauthorrefmark{2}aman.verma.nitrr@gmail.com,
\IEEEauthorrefmark{3}raghavendra.ramachandra@ntnu.no}}

\markboth{Journal of \LaTeX\ Class Files,~Vol.~14, No.~8, August~2021}%
{Shell \MakeLowercase{\textit{et al.}}: A Sample Article Using IEEEtran.cls for IEEE Journals}


\maketitle

\begin{abstract}
An iris presentation attack detection (IPAD) is essential for securing the personal identity in widely used iris recognition systems. However, the existing IPAD algorithms do not generalize well to unseen and cross-domain scenarios because of capture in unconstrained environments and high-visual correlation amongst bonafide and attack samples. These similarities in intricate textural and morphological patterns of iris ocular images contribute further to performance degradation. To alleviate these shortcomings, this paper proposes DFCANet – Dense Feature Calibration and Attention Guided Network which calibrates the locally spread iris patterns with the globally located ones. Uplifting advantages from feature calibration convolution and residual learning, DFCANet generates domain-specific iris feature representations. Since some channels in the calibrated feature maps contain more prominent information, we capitalize discriminative feature learning across the channels through channel attention mechanism. In order to intensify the challenge for our proposed model, we make DFCANet operate over non-segmented and non-normalized ocular iris images. Extensive experimentation conducted over challenging cross-domain and intra-domain scenarios highlight consistent outperforming results. Compared to state-of-the-art methods, DFCANet achieves significant gains in performance for the benchmark IIITD-CLI, IIIT-CSD and NDCLD’13 databases respectively. Further, a novel incremental learning based methodology has been introduced so as to overcome disentangled iris-data characteristics and data-scarcity.  This paper also pursues the challenging scenario that considers soft-lens under attack category with evaluation performed under various cross-domain protocols. The code will be made publicly available. 
\end{abstract}


\section{Introduction}

\label{sec1}
\begin{figure}[t]
	\begin{center}
		\includegraphics[width=1\linewidth]{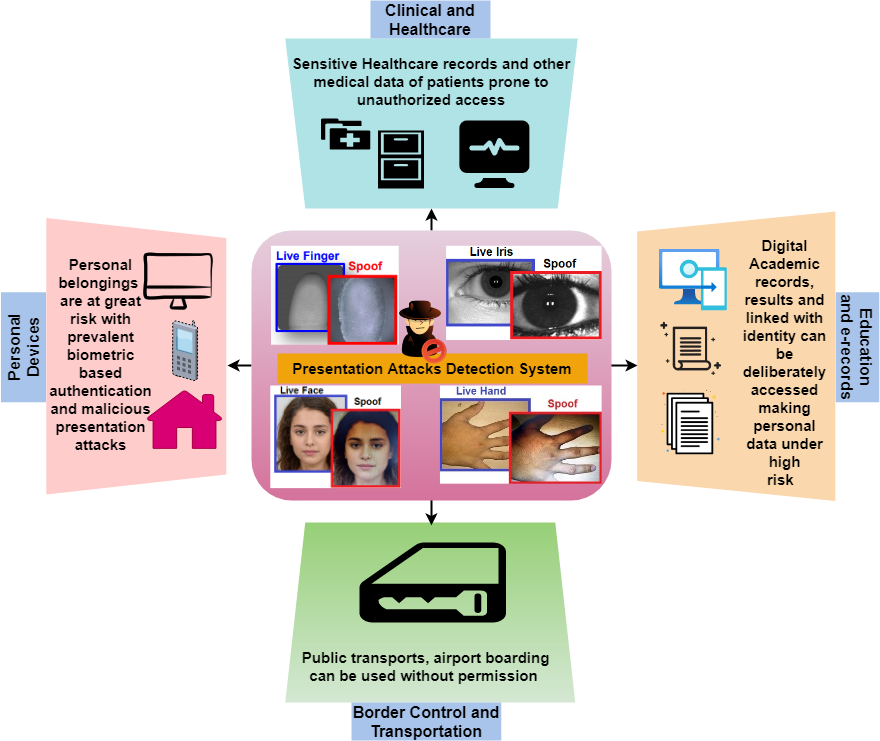}
	\end{center}
	\caption{Presentation attack detection and its applications for augmenting digital revolution via robust user authentication}
	\label{fig:2}
\end{figure}

The fourth industrial revolution has been representing the transition to digitization of socio-commercial activities, impacting almost every sector of human life. It has also significantly emphasized personalization and at the same time raising questions on identity measures and security awareness \cite{jain201650}. From smart homes to autonomous vehicles, every cyber physical system is now being governed by the user’s identity \cite{garg2019toward}. Smartphones are another integral part of this smart-cybernetical infrastructure as they have become frequent storage medium for personal information such as financial documents, credentials, health-records and other multimedia \cite{prabhakar2003biometric, jaswal2016knuckle}. The access medium to these facilities has been the personal identity. Most of the smartphone are equipped with fingerprint/ face/iris/ hand identification technology to login or to pay in mobile applications \cite{jaswal2016knuckle, jain201650}. On the same note, the healthcare industry, medical companies and smart cities have started to adopt an array of security systems that use data from a patient’s own biological features to protect their medical records. Such factors critically emphasize the requirement of an impregnable identity-preservation measures. Biometric traits \cite{jain2021biometrics}, both physical and behavioral, have proven effective in many large-scale applications such as national identification, smartphone unlock, border control, payment apps etc.
However, the use of authentication mechanisms based on biometric solutions can often collide with national and international regulations which tend to protect individual privacy and sensitive information \cite{drozdowski2020demographic}. While the use of biometrics for security can be developed in full awareness of the users, in case of safety, the collection of biometric data concerns a more subtle concept of privacy. With the increased deployment in private and public context, the security of biometric systems against malicious attacks becomes critical. One such attack that is being increasingly gained attention is the presentation attack (PA), where a sample is presented to the sensor with a fake or deliberately altered biometric trait in order to fool the system \cite{uludag2004attacks}. Presentation attacks may be carried out to impersonate an identity during verification; to conceal an identity during recognition or to create a virtual identity during enrollment \cite{tolosana2019biometric}. Solutions to detect these attacks are referred to as Presentation Attack Detection (PAD) biometric system which can determine whether a sample presented to a sensor is from a bonafide one or is a spoof presentation. Figure \ref{fig:2} is focusing on necessity of generalize PAD algorithms in unconstrained operating conditions and more realistic scenarios for unbreach user authentication.  

\subsection{Problem statement and open challenges}

 Compared with face or fingerprint recognition in terms of accuracy, iris has been considered as one of the most stable, accurate and reliable biometric authentication technologies \cite{daugman2003importance, daugman2007new}, hence research in iris presentation attack detection (IPAD) has received substantial attention. The possible forms of iris presentation attacks include cosmetic contact lenses, holographic eyes, textured contact lenses, prosthetic eyes, printed iris images, iris videos, paper iris printouts, drug-induced iris manipulation, and fake eyeballs \cite{morales2019introduction, marcel2019handbook, fang2021cross}. LivDet-Iris is an international competition series launched in 2013 to assess the current state-of-the-art in iris PAD by the independent evaluation of algorithms \cite{marcel2019handbook, das2020iris}. In addition, results from the LivDet-Iris 2017 competition illustrated that state-of-the-art methods still obtain limited accuracy in spoof detection \cite{yambay2017livdet}. This was presented specifically for unseen-attack and cross-dataset scenarios, where characteristics of testing samples differ from the samples considered for training. Under these stringent evaluation protocols, generalization of the I-PAD systems have been observed to get affected \cite{gupta2014iris, fang2021cross}. \par 
 IPAD systems are subject to multitude of challenges and specifically with respect to ocular iris images. These images are characterized by complex iris patterns (some example images are shown in input of Figure \ref{fig:2b}) as well as morphological artefacts included in annular structure of the eye such as eye-lashes.  Another challenging aspect of ocular iris images is the fact that the iris biometric template remains concealed within limited boundaries in the entire image. Uneven lighting conditions and occluded image capture further degrades attention to important details. Moreover, segmentation of iris in the ‘wild’ is an intricate task because of pose and lighting conditions-based artefacts. The above-forth problems deepen when I-PAD algorithms are evaluated under unseen and cross-domain scenarios. Further, it is observed that soft-lens bears high visual similarity with the normal iris images and ironically this aspect has been laid very limited attention in existing works \cite{yadav2018fusion, fang2021iris}.
 Therefore, there is a dire requirement of an ideal IPAD algorithm that enforces robust decision boundaries between all forms of bonafide and presentation attacks while generalizing well under unseen and cross-domain scenarios.

\subsection{Our contribution} To address the aforementioned problems, we present an end-to-end deep learning based I-PAD framework- DFCANet whose detailed architecture is depicted in Figure 2.  In light of this, the following main aspects are considered: 
\begin{itemize}
   \item how to design a deep-learning based framework for IPAD that operates in the domain-specific paradigm? 
 \item how to achieve a consistent IPAD performance for challenging cross-domain scenarios wherein drastic drifts between training and testing samples is present? 
 \item how to alleviate performance against data-scarce, un-constrained and zero-preprocessing conditions?
 \end{itemize}
In essence of catering the above-mentioned points, our proposed model DFCANet uplifts advantages of domain-enriched and multi-scale feature learning, and thus generalizing well for unseen and cross-domain IPAD scenarios. Specifically, our model operates over raw iris images without applying any pre-processing and then it passes to feature calibration assisted backbone network for robust presentation attack detection. In order to validate potential in our model, we conduct challenging experiments and extensive ablation studies on three benchmark datasets viz IIITD-CSD, NDCLD-13, IIITD-CLI and obtain significant improvement in performance over state-of-the-art results. We present the contribution of our work in four-folds and validate the effectiveness through extensive ablation studies:
\begin{itemize}
    \item IFCNet (shown in Figure \ref{fig:2a}): We exploit the utilization of residually connected iris feature calibration convolutions that enable more domain specific knowledge in terms of local-global patterns present in the iris-template to be transferred for robust discrimination between bonafide and attack ocular iris images.
    \item CAM (shown in Figure \ref{fig:2b}): We employ a simple yet effective channel attention module so as to enhance the discriminative power of local-global features across the channels.
    \item To the best of the author's knowledge, this is the first IPAD approach which consider soft lens as both bonafide and attack under various experiments.
    \item We present a novel incremental learning methodology to augment the performance on challenging characteristics of the NDCLD-13 dataset.
\end{itemize}

Our work advances the state-of-the-art that has received very little attention in IPAD literature by considering the raw ocular iris images under challenging cross domain and soft-lens as attack scenarios.\par

The rest of the paper is structured into four main sections. Section II critically reviews the existing works on IPAD  with an emphasis over deep learning based methods. Section III presents the methodology of our approach and includes the details of the network and training considered in this work. Section IV details about datasets, experiments, and testing protocol employed for performance evaluation. Key findings and discussions are summarized in the last section.

\section{Related Work}
In recent years, many studies on iris biometrics have started to employ deep learning schemes \cite{sharma2020d, kohli2016detecting, kuehlkamp2018ensemble} and presented remarkable progress in IPAD performance. Though hand crafted features have also shown significant progress in IPAD, specifically for intra-database testing scenarios. However, their progress is still far from satisfactory to new application scenarios. As a result, the handcrafted features reflect limited aspects of the problem, yielding a detection accuracy that is low and varies with the characteristics of presentation attack iris images. The first iris PAD approach \cite{daugman2000wavelet} was probably proposed in early 2000 and Fourier analysis was applied to detect artificial patterns in printed contact lenses presented to a sensor. The other well-known hand-crafted features that have been used in IPAD include hierarchical visual codebook \cite{sun2013iris}, local binary pattern \cite{he2009efficient}, weighted local binary pattern \cite{zhang2010contact}, spatial pyramidal matching \cite{hu2016iris}, bag of words \cite{gragnaniello2016using} etc. After that, several  machine learning based PAD algorithms are developed which yield high
accuracy in constrained environments. In \cite{raja2015video}, authors made efforts to enhance the subtle phase information in the eye for detecting iris video presentation attacks in visible spectrum. In \cite{raghavendra2015robust}, authors presented an unsupervised machine learning approach to detect vulnerability of iris presentation attacks. In contrast, IPAD based on deep learning approaches \cite{menotti2015deep,fang2021iris, chenexplainable} have resulted into impressive accuracy's due to their ability to extract highly domain specific iris representations. The first work that proposed a deep architecture for IPAD called as SpoofNet\cite{menotti2015deep}. Authors developed a deep framework built for iris liveness detection utilizing triplet convolutional networks\cite{pala2017iris}.
Moreover, some of the schemes explored combination of hand crafted features with deep learning and achieved good results \cite{yadav2018fusion, kuehlkamp2018ensemble}. Unlike fusion of hand crafted and CNN features, authors in \cite{fang2020deep} presented a multi-layer deep fusion scheme extracting different level of information from multiple layers of the network. Apart from fusion schemes, most recently, attention based deep learning framework named pixel wise binary supervision network \cite{fang2021iris} was given to capture fine grained pixel level information that can be emphasized for making accurate IPAD decision. Likewise in \cite{chenexplainable}, authors presented an explainable attention-guided IPAD that can improve both the generalization and explanation capability of existing approaches. Authors in \cite{yadav2019detecting}, presented DensePAD method to detect presentation attacks by utilizing DenseNet-121 architecture. Likewise, authors in \cite{sharma2020d}, also exploited the architectural benefits of DenseNet to propose an IPAD scheme that tested on four different sub-datasets of LivDet-Iris 2017
databases. Although their method achieved good results on LivDet-Iris 2017 databases, however, the performance dropped in the case of cross-database scenarios. In recent work\cite{gupta2021generalized}, authors presented a generalize IPAD to resolve degrade in performance of DNN's against unseen database, unseen sensor,
and different imaging conditions. Some other studies were developed to create synthetic iris-like
patterns using generative adversarial network \cite{kohli2017synthetic} with applications to IPAD. However, these methods are not scalable to more than
two domains and often show instability in the presence of multiple domains. \cite{yadav2021cit} proposed domain invariant styling n/w to generate good quality iris synthetic images to assist training the IPAD models for better performance. Some recent studies have been reporting survey papers \cite{czajka2018presentation, morales2019introduction, marcel2019handbook}, new datasets and conducting Iris-PAD competitions \cite{das2020iris, yambay2017livdet, yambay2019review} that has increased the potential of DNN based IPAD approaches under constrained environments. However, the new DNN based IPAD algorithms are still need to be adapted to various cross-domain settings plus robust to new attacks so that future IPAD systems must be generalized against various cross-domain and open-set conditions.

\section{Methodology}

\begin{figure*}[t]
	\begin{center}
		\includegraphics[width=1\linewidth]{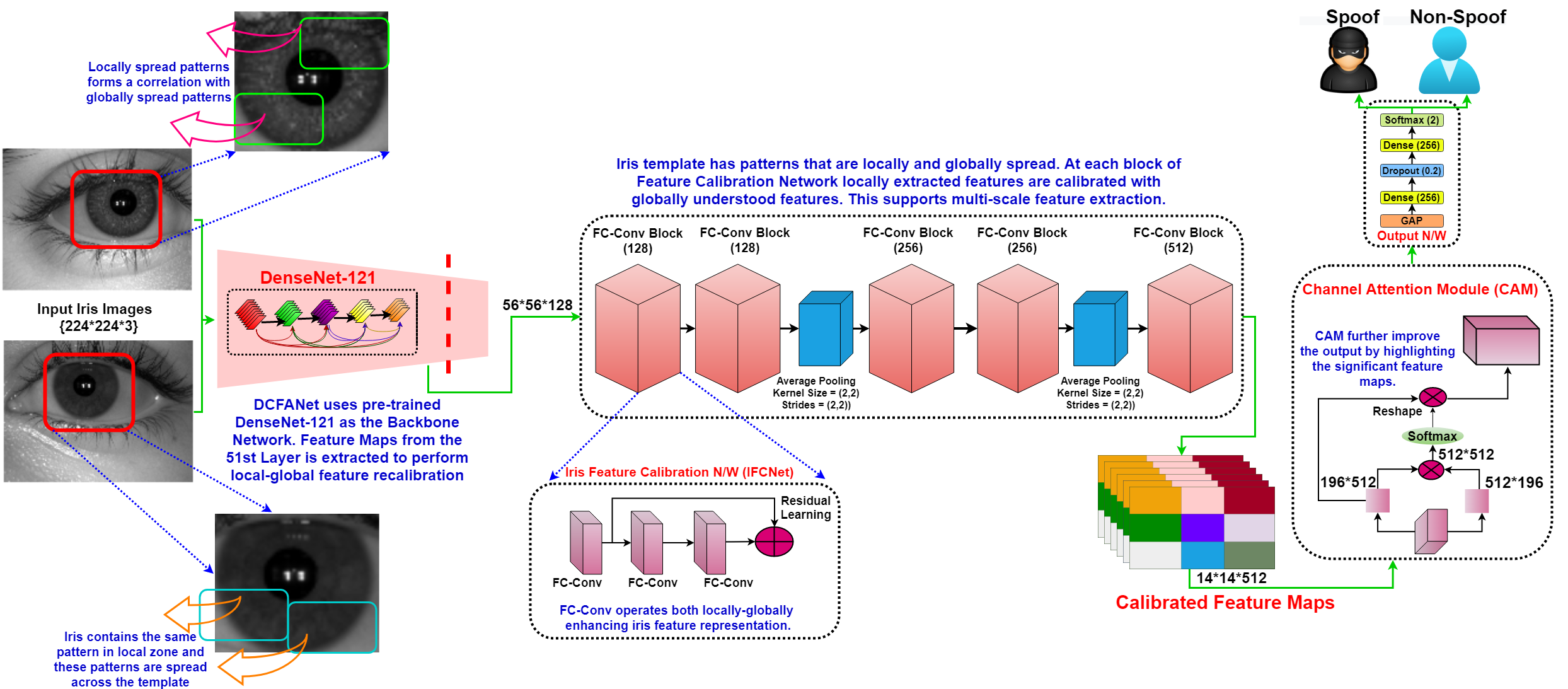}
	\end{center}
	\caption{The proposed DFCANet is composed of majorly three blocks, namely DenseNet121 backbone network, iris feature calibration network (IFCNet) and channel attention module (CAM). The feature maps extracted from this pipeline is used for classifying iris-based presentation attacks.}
	\label{fig:2b}
\end{figure*}

In this section, we describe the proposed IPAD model, namely DFCANet (shown in Figure \ref{fig:2b}), by focusing on how the network can be instrumental for addressing the open-research problems in IPAD. In particular, by analysing intrinsic iris patterns, we formulate two key postulates (see Figure 2) with respect to IPAD problems : i) Iris contains the same pattern in local-neighborhoods and these patterns are further spread across the entire iris structure. ii) These locally spread patterns form a correlation with each other and hence at a global scale they constitute a homogeneous pattern. DFCANet’s architectural design gains motivation from these postulates to improvise over generalization in unseen and cross-domain settings. Aiming towards the same, DFCANet does not operate over ROI-segmented or Iris-Normalized images but rather inputs are rescaled images of dimensions $R^{(224\times224\times3)}$. Consequently, the rescaled images are given as input to the proposed model which encompasses 3 major components namely DenseNet121, Iris Feature-Calibration Network (IFCNet), Channel Attention Module (CAM). Each of those components are explained in the subsequent subsections.

\subsection{Backbone Network}
DenseNet-121~\cite{huang2017densely} has been used as the backbone network for DFCANet. While the DenseNet121 was kept fine-tunable, the weights for initialization have been adapted from ImageNet~\cite{krizhevsky2017imagenet} pretraining. Without applying any pre-processing such as segmentation and normalization, the original iris images having input dimension $224\times224\times3$ are inputted to the DenseNet. The extracted features from DenseNet form a rich representation attributed to dense connections and feature-reuse. This enhances the gradient flow during learning and enables feature-maps from initial layers to interact with the ones at later layers for multi-scale feature extraction. With enhanced feature-flow locally learnt patterns at lower layers are robustly correlated with the globally located representations at higher layers benefitting presentation attack detection system. For attacks such as print-attacks and scan-attacks, overall textural patterns are the key-distinguishing factor.  Since, the DenseNet has an enlarged receptive-field because of feature-flow and deeper architecture, it is also instrumental in understanding the globally spread textures. Hence, it is robust for scanner-based attacks also. In DFCANet, we took out the features from the $51^{st}$ conv layer, resulting into the output dimensions of $R^{(56\times 56\times 128)}$.  

\subsection{Iris Feature Calibration Network: IFCNet}
\begin{figure}[t]
	\begin{center}
		\includegraphics[width=1\linewidth]{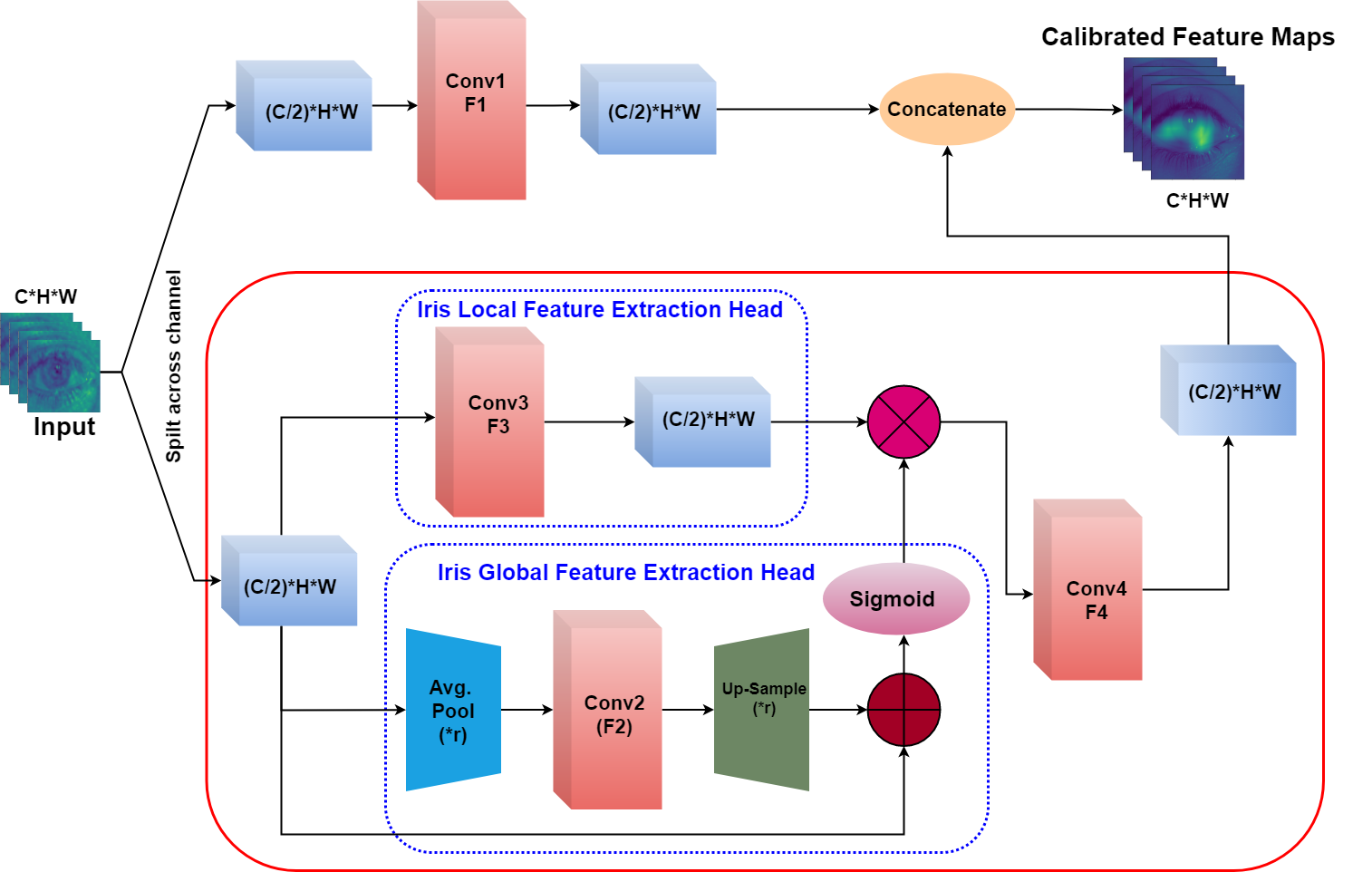}
	\end{center}
	\caption{IFCNet: Iris feature calibration convolution for learning of enhanced iris feature representation}
	\label{fig:2a}
\end{figure}

For modelling iris-patterns it is essential to correlate the locally activated nuances with the globally understood features as similar textural patterns exist in the entire iris template. To this end, DFCANet utilizes a novel network – Iris Feature Calibration Network (IFCNet) to harness local-global feature correspondence. The output of the base-network Densenet-121 serves as the input to the IFCNet of the DCFANet. Architecturally, IFCNet is a stack of 5 FC-Blocks and in-turn a FC-Block comprises of three residually connected Feature Calibration Convolutions (FC-Conv). FC-Conv \cite{liu2020improving} has its functionality in helping the model learn information at local-global scales both individually as well as jointly. Further, it also helps in tuning the features extracted in a region restricted with kernel-size, in accordance with the global feature hierarchy formed, leading to better domain-specific features. Architecture of the FC-Conv has been illustrated in Figure \ref{fig:2a}.

\textbf{Working Operation:} First, the input $I \epsilon R^{H \times W\times C}$ is split across the channel dimension into two with number of channels in each of the resultant being a hyperparameter, for the case of DFCANet the input was divided equally into two halves. Height, Width and Channel dimensions are respectively represented by H, W and C respectively. Let, $I_{1}$ be the first half of the channel-wise split while $I_{2}$ is the second half. The first half operates through a regular 2D-Convolution, while the second half is made to go through the feature calibration head which is constituted of two subheads: local feature extraction head (LFEH), and global feature extraction head (GFEH). Let $F_{1}$represent the first convolution, thus, it can be expressed as:  

\begin{equation}
    I_{1}^{'} = F_{1}(I_{1}); 
    I_{1}^{'} \epsilon R^{H \times W \times C/2} 
\end{equation}
In LFEH, the input is operated using a regular 2D-Convolution $F_{3}$, we refer this as local convolution because of its localized receptive field and can be expressed as:
\begin{equation}
    I_{2-Local}^{'} = F_{3}(I_{2});I_{2-Local}^{'} \epsilon R^{H \times W \times C/2} 
\end{equation}

Whereas in the GFEH, the input is down-sampled using Average-Pooling operation, which is succeeded by a 2D-Convolution ($F_{4}$) and a bilinear up-sampling reduction in spatial-dimension with the average pooling lets the model develop representations on the basis of globally compressed information. Let, $Up(.)$ represent the bilinear up-sampling and $Avg.Pool()$ represent the 2D average pooling. Then, it can be expressed as: 

\begin{equation}
    I_{2-Global}^{'} = Up(F_{2}(Avg.Pool(I_{2}))); I_{2-Local}^{'} \epsilon R^{H \times W \times C/2} 
\end{equation}

The output of the upsampling is added with the input to the Calibration head, this ensures stability in flow of gradients as well as highlights important features. The generated output is activated through a sigmoid operator so as to undermine the importance of the globally understood features and finally in the feature calibration step, this output is multiplied elementwise with the output of the LFEH. Mathematically, it can be expressed as:

\begin{equation}
    I_{2}^{'} = F_{4}(I_{2-Local}^{'}\otimes \sigma (I_{2-Global}^{'} + I_{2}));
    I_{2-Local}^{'} \epsilon R^{H \times W \times C/2} 
\end{equation}

This operation emphasizes the important local features with respect to global feature representation. This output is subjected to another 2D Convolution ($F_{4}$) so as to generate the final output ($Y_{FC}$) of the Calibration Head. The FC-Block is the concatenation of $I_{1}^{'}$ and the output of Feature-Calibration Head $I_{2}^{'}$. Mathematically, it can be expressed as:
\begin{equation}
  Y_{FC} = I_{1}^{'} \parallel I_{2}^{'}; Y_{FC}\epsilon R^{H \times W \times C} 
\end{equation}

Here, $\sigma$ and  $\otimes$ represents element-wise sigmoid operation and multiplication respectively. Likewise, $\parallel$ represents channel-wise concatenation of two-tensors. The convolutions $F_{3}$ and $F_{2}$ are ReLU activated while the convolutions $F_{1}$  and $F_{4}$ are linearly activated but are succeeded by batch normalization and ReLU activation. Convolution operation $F_{1}$, $F_{3}$ and $F_{4}$ adhere to same kernel-size ($k_{1}$) while because of reduced dimension Convolution $F_{2}$ has a smaller kernel-size $k_{2}$. Accredited to limited number of channels per kernel the parameter efficiency of FC-Conv is significantly higher than regular 2D Convolutions. Owing to this fact, FC-Conv allows to stack-up multiple layers of it, hence forming a network robust enough to capture highly non-linear relationships.  

\begin{small}
\begin{table}[ht]\scriptsize
\caption{Description of different parameters of FCNet}
	\label{tab:1}
  \begin{center}
\begin{tabular}{|l|l|l|l|l|l|}
\hline
Model                                                                                      & Sub-module & k1    & k2    & \begin{tabular}[c]{@{}l@{}}Avg. \\ Pooling\end{tabular} & \begin{tabular}[c]{@{}l@{}} o/p\\  channels\end{tabular} \\ \hline
{FC Block1}                                                                 & FC Conv1   & (3,3) & (7,7) & (11,11)                                                 & 128                                                       \\ \cline{2-6} 
                                                                                           & FC Conv2   & (3,3) & (7,7) & (11,11)                                                 & 128                                                       \\ \cline{2-6} 
                                                                                           & FC Conv3   & (3,3) & (7,7) & (11,11)                                                 & 128                                                       \\ \hline
{FC Block2}                                                                 & FC Conv1   & (3,3) & (7,7) & (11,11)                                                 & 128                                                       \\ \cline{2-6} 
                                                                                           & FC Conv2   & (3,3) & (7,7) & (11,11)                                                 & 128                                                       \\ \cline{2-6} 
                                                                                           & FC Conv3   & (3,3) & (7,7) & (11,11)                                                 & 128                                                       \\ \hline
\begin{tabular}[c]{@{}l@{}}Avg. Pool2D\\ Pool Size = (2,2);\\ Strides = (2,2)\end{tabular} & -          & -     & -     & -                                                       & 256                                                       \\ \hline
(1×1) Conv2D                                                                               & -          & -     & -     & -                                                       & 256                                                       \\ \hline
{FC Block3}                                                                 & FC Conv1   & (3,3) & (5,5) & (9,9)                                                   & 256                                                       \\ \cline{2-6} 
                                                                                           & FC Conv2   & (3,3) & (5,5) & (9,9)                                                   & 256                                                       \\ \cline{2-6} 
                                                                                           & FC Conv3   & (3,3) & (5,5) & (9,9)                                                   & 256                                                       \\ \hline
{FC Block4}                                                                 & FC Conv1   & (3,3) & (5,5) & (9,9)                                                   & 256                                                       \\ \cline{2-6} 
                                                                                           & FC Conv2   & (3,3) & (5,5) & (9,9)                                                   & 256                                                       \\ \cline{2-6} 
                                                                                           & FC Conv3   & (3,3) & (5,5) & (9,9)                                                   & 256                                                       \\ \hline
\begin{tabular}[c]{@{}l@{}}Avg. Pool2D\\ Pool Size = (2,2);\\ Strides = (2,2)\end{tabular} & -          & -     & -     & -                                                       & -                                                         \\ \hline
(1×1) Conv2D                                                                               & -          & -     & -     & -                                                       & 512                                                       \\ \hline
{FC-Block4}                                                                 & FC Conv1   & (3,3) & (3,3) & (7,7)                                                   & 512                                                       \\ \cline{2-6} 
                                                                                           & FC Conv2   & (3,3) & (3,3) & (7,7)                                                   & 512                                                       \\ \cline{2-6} 
                                                                                           & FC Conv3   & (3,3) & (3,3) & (7,7)                                                   & 512                                                       \\ \hline
\end{tabular}
  \end{center}
\end{table}
\end{small}

\textbf{Network Design:} IFCNet’s architecture is designed in a pyramidal fashion. It’s module FC-Block consists of three FC-Conv layers, each having the same hyperparameters. The output of the first FC-Conv is residually added with the output of the final FC-Conv. This residual has two-fold benefits – Firstly, there is feature flow which enhances global-level representations and secondly, it helps in tackling vanishing gradients for the deep architecture of DFCANet. The overall architectural-design of IFCNet has $5$ FC-Blocks; after second and the fourth FC-Block there is average pooling operation which is followed by a 2D- Convolution of kernel-size $1\times 1$ and output filters equivalent to twice of the previous layer. The detailed parametrization of IFCNet is given in Table \ref{tab:1}. 

\textbf{Justification :} Biometric identity in the iris comes from its specific patterns, texture features, and its morphology; moreover, these features are locally distributed as well as have global importance. The traditional feature extraction methods used in iris PAD~\cite{kohli2016detecting} extracts a specific type of feature at once thus they are not capable of forming a correlated representation between multiple types of features leading to degraded information. While deep learning based features are extracted in correlated manner, the residual or multi-scale features~\cite{sharma2020d} in deeper networks do not pay attention to important locations and thereby arising need of more discriminative features; Even Positional attention~\cite{chenexplainable} based features are deprived of forming representations which can collect local as well global prominence of features. In turn FC-Block calibrates locally extracted features with globally understood features, further this can be thought as an attention operation in spatial domain which is done at both local and global scales and thus improving the receptive field of the model on an overall basis. The final advantage of this sort of feature extraction is that the generalization of cross-domain attacks is improved.

\subsection{Channel Attention and O/P Layer N/W}
Once the features are calibrated through the FCNet, it is of utmost importance to give higher significance to essential channels. Conversely, suppression of trivial-information present in various channels is necessitous for two-fold purpose: i) Enhancement of feature-discriminability, ii) Stabilization of training. To solve the same purpose, CAM[4] has been used and it has been illustrated in Figure 2. CAM highlights the important channels yet alleviating a zero-parametric design. The operation flow of the CAM is explained as follows:-  CAM takes in input $X \epsilon R ^ {H \times W \times C} $ where H, W, and C denote height, width and channel dimensions in the input tensor X, respectively. The input is then reshaped to $Q \epsilon R ^ {C \times N}$ with N being equal to $H\times W$. $Q$ and $Q^{T}$ i.e., transpose of $Q$ are multiplied to get 2D Matrix which is activated through Softmax function so as to get $U \epsilon R ^ {C \times C}$. Each of the elements $U_{ij}$ is weight measuring the importance of channel $‘j’$ with respect to channel $‘i’$. Following this, $U$ is multiplied with $Q$ and the corresponding output $I_{Refined}$  is reshaped to $\epsilon R ^ {H \times W \times C }$ . $I_{Refined}$ are the refined feature maps which have been generated as a result of  original feature multiplication with corresponding attention weights. The output of the CAM is $I_{CAM}$ which is weighted sum of $I_{Refined}$ with $X$ with the weighting factor $\beta$ chosen 1 for DFCANet but in general being $\beta \epsilon [0,1]$. Mathematically, it can be expressed as:

\begin{equation}
    I_{CAM} = \beta \times I_{Refined} + X
\end{equation}

\begin{figure}[t]
	\begin{center}
		\includegraphics[width=1\linewidth]{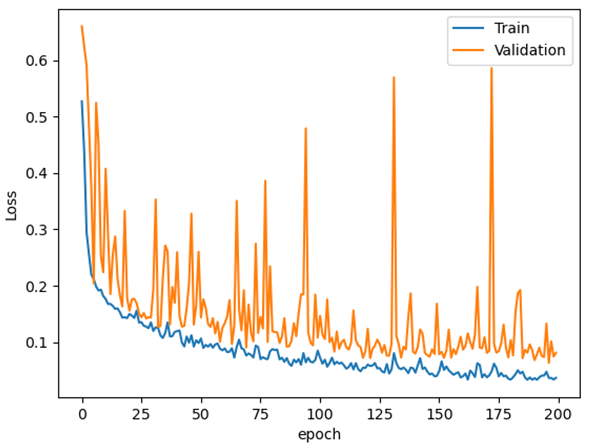}
	\end{center}
	\caption{Training and Validation Loss Curves}
	\label{fig:3a}
\end{figure}

\subsection{Output Network and Training Details}
The iris-images resized to dimensions $R ^ {224 \times 224 \times 3}$, is given as input to the DFCANet. These images are augmented with random perturbations including spatial shifting, shearing, enhances the representation by giving more weights to channels with prominent information. After CAM, global average pooling (GAP) is applied which is followed by a dense layer with 256 neurons. Further, to add regularization, dropout is applied with dropout-rate being $0.2$. Following the dropout, dense layer having $256$ neurons is applied. This is the penultimate layer and after this final layer is present which is sigmoid activated. The final embeddings are extracted and given to binary cross entropy loss. While backpropagation, adam optimizer with uniform learning rate $0.0001$ is utilized. Here, the hyperparameters $\beta_{1} = 0.9$ and $\beta_{1} = 0.999$. Every model is trained for $200$ epochs with batch-size being $32$. During training for $200$ epochs, the model achieving maximum validation accuracy is chosen and is saved. Figure \ref{fig:3a} illustrates the loss optimization for both training and validation set. From the curve it is evident that convergence of the model occurs around about $150$ epochs but there were subtle developments when the model is trained for $200$ epochs. During the course of training, model achieving the highest validation accuracy is saved. The proposed framework has been implemented using TensorFlow 2.0 while all the experimentation has been conducted in NVIDIA’s RTX-3090 GPUs.

\section{Experiments and discussions}
In this section, extensive IPAD experiments and their related contents are demonstrated. First, we introduce the used IPAD datasets, containing three most commonly used IPAD datasets. The evaluation metrics are also introduced in this part. To validate the potential in approach, we have conducted experiments under challenging strategies including i) cross-sensor ii) cross-database iii) intra-sensor and iii) combined sensor. Further, to demonstrate the effectiveness of the FCNet and CAM structures independently, the ablation experiments are conducted and compared with the proposed framework.  Comparisons with other state-of-the-art methods are also conducted which demonstrate its excellent performance. In addition, we introduce baseline experiments based on incremental learning and performed cross database experiments as well to illustrate how practically DFCANet in solving IPAD problems.

\subsection{Datasets, Testing Protocols, and Performance Metrics}
The proposed framework is evaluated on three databases comprising of various types of presentation attacks captured by different sensors. Table  tabulates brief description about database in terms of quantities of data, sensor type and their respective characteristics. 

\textbf{ 1) IIITD-CLI \cite{yadav2014unraveling} and testing protocol:} This dataset is composed dual-eye ocular images collected from 101 subjects and there are total of 6750 iris images. Two eye-scanners: cogent and vista F2AE have been employed for the task. There are contact-lens based attacks in the dataset i.e., Soft and Textured Lens. We have considered a challenging strategy of taking Soft-Lens as an attack. The following instance is challenging as it gives very-similar appearance to normal iris image. This scenario was taken into account because of two-fold reasons: i) We wanted our proposed model to be invariant to any of the occlusions or hindrance ii) There is a high possibility that soft-lens can be as a spoofing medium. The lens attacks are with 4 different colors with the lenses being from the makes CIBA Vision, Bausch and Lomb. For conducting the experiments with this database, 50-50\% data was split for training and testing in hold-out fashion.

\textbf{2) NDCLD’13 \cite{doyle2013variation} and testing protocol:} This database comprises of two datasets ND-I and ND-II collected from two different sensors, AD100 IrisGaurd and LG4000 respectively. Similar to IIIT-CLI, this dataset comprises of normal iris images and corresponding soft, textured contact lens images. Keeping consistent with the strategy, attacks that have been considered from this dataset are textured contact lens and soft contact lens with the lenses coming from CIBA Vision, Johnson-Johnson and Cooper-Vision. Both ND-I and ND-II have been provided with a training and testing set. ND-I consists of 600 images in training while 300 in testing. For the case of ND-II, there are 3000 images given under training and 1200 for testing.  
 
\textbf{3) IIITD-CSD \cite{kohli2016detecting} and testing protocol:} In accordance with the training-testing strategy used in IIITD-CLI and NDCLD’13 datasets, IIITD-CSD consists of print, scan, textured and soft contact lens-based attacks. It is the biggest dataset considered in this study. In total it consists of 17036 ocular-iris images.  For eye-scan, Cogent and Vista F2AE sensors had been utilized and while for print and scan attacks HP Flatbed Optical Scanner and Cogent-CIS 202 was used. Experimentation with this database involved 50-50\% splitting for training and testing set in hold-out fashion.

\par To fairly evaluate the performance
of different methods, all experiments have been validated with a set of distinguishing metrics such as Attack Presentation Classification Error Rate (APCER), Normal Presentation Classification Error Rate (NPCER), and Average Classification Accuracy (AA).

\begin{small}
\begin{table}[ht]\scriptsize
\caption{Testing Performance on intra-sensor experiments in terms of AA (\%), APCER (\%), NPCER (\%), and ACER (\%) }
	\label{tab:2}
  \begin{center}\begin{tabular}{|l|l|l|l|l|l|l|}
\hline
Datasets                   & Training & Testing & AA  & APCER  & NPCER  & ACER \\ \hline
{IIITD-CLD} & Cogent               & Cogent              & 99.33              & 2.03       & 0.31       & 1.17      \\
\cline{2-7} 
                           & Vista                & Vista               & 99.80              & 0.40       & 0.17       & 0.27      \\
                           \hline
{IIITD-CLI} & Cogent               & Cogent              & 98.90              & 1.52       & 0.87       & 1.19      \\
\cline{2-7} 
                           & Vista                & Vista               & 99.79              & 0.41       & 0.90       & 0.25      \\
                               \hline
{NDCLD13}   & LG4000               & LG4000              & 95.91              & 9.75       & 1.25       & 5.50      \\
\cline{2-7} 
                           & AD100                & AD100               & 76.00              & 44.00      & 14.00      & 29.00 \\
                            \hline
\end{tabular}
  \end{center}
\end{table}
\end{small}

\subsection{Experimental Results and Discussions}
IPAD systems are often prone to performance degradation in unconstrained settings. Hence, to provide a stringent validation of the proposed DFCANet based IPAD system, we have performed cross-domain experiments. Results and corresponding discussions of the same has been presented using the detection error trade-off (DET) curve with equal error rates (EER) and other performance metrics discussed in previous sections.

\begin{figure}[t]
	\begin{center}
		\includegraphics[width=1\linewidth]{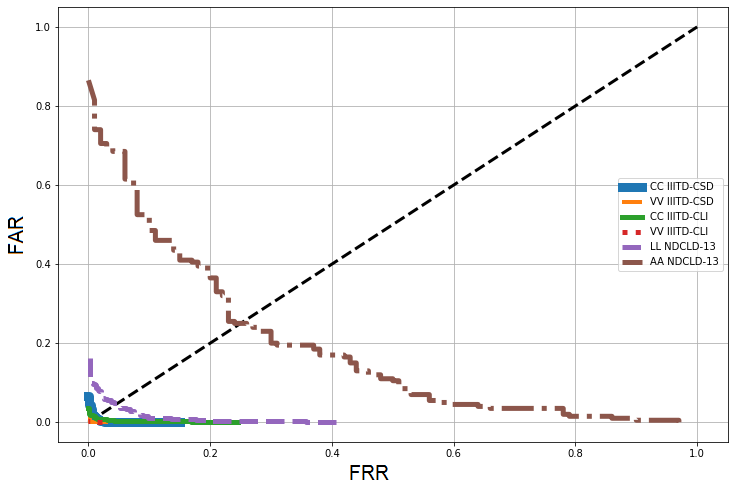}
	\end{center}
	\caption{DET plot for comparing different intra-sensor experiments}
	\label{fig:det1}
\end{figure}

\textbf{1) Intra-Sensor Experiments:} Intra-Sensor experiment involved training the model on data from a sensor, while testing it on the data from the same sensor. As discussed earlier, for the case of IIITD-CSD and IIITD-CLI, 50-50\% hold-out data division was considered while NDCLD’13 dataset provided the training and testing data individually. Results obtained under this setting has been tabulated in Table \ref{tab:2} . The proposed DFCANet attains significant performance in all the three databases. For the Vista-trained and  Vista-tested model on IIITD-CLI database, ACER obtained is 0.25\% and is the lowest amongst all the models under this experiment. A comparative performance is observed by the Vista-trained and Vista-tested model of the IIITD-CSD database. In NDCLD’13 database, the model trained-tested on LG4000 dataset, the obtained average accuracy of 95.91\% is significant, but is the model is comparatively less robust in terms of classifying of bonafide attempts. In-fact, it is evident that when the proposed model is trained and tested on similar sensor, there is low APCER but lower NPCER i.e., the model is more robust for detecting spoofs. Figure \ref{fig:det1} illustrates the  DET curves for comparing performance of intra-sensor based experimental results. From the mentioned observation it can be inferred that the decision boundaries learnt by the model is more spanning in the attack-class’s region, a possible reason for the same can be multi-cluster and multi-center data formations of the attack class. The average accuracy attained by the model is highest for the IIITD-CSD, specifically for Vista sensor achieving 99.80\% of the times correct detections. It can also be concluded from the table that the model is better for Cogent and Vista sensors, when compared with LG4000 and AD100. A specific reason for the same can be occlusions and sensing inconsistencies of the ocular-iris images in the NDCLD’13 dataset. Particularly, the data from AD100 sensor comprises of just 600 images, and because of insufficient training quantities the architectural depth of the model limits itself in capturing non-linearities leading to sub-optimal convergence.

\begin{small}
\begin{table}[ht]\scriptsize
\caption{Testing Performance on inter-sensor experiments in terms of AA (\%), APCER (\%), NPCER (\%), and ACER (\%) }
	\label{tab:3}
\begin{center}\begin{tabular}{|l|l|l|l|l|l|l|}
\hline
Datasets                   & Training & Testing  & AA & APCER  & NPCER & ACER  \\ \hline
IIITD-CLD & Cogent               & Vista               & 99.33             & 0.41      & 0.72      & 0.56     \\ \cline{2-7} 
                           & Vista                & Cogent              & 96.49             & 13.98     & 0.87      & 7.42     \\ \hline
IIITD-CLI & Cogent               & Vista               & 99.19             & 0.31      & 1.04      & 0.67     \\ \cline{2-7} 
                           & Vista                & Cogent              & 95.52             & 10.10     & 1.64      & 5.87     \\ \hline
NDCLD13  & LG4000               & AD100               & 90.22             & 13.00     & 8.16      & 10.58    \\ \cline{2-7} 
                           & AD100                & LG4000              & 85.88             & 25.74     & 8.20      & 16.97    \\ \hline
\end{tabular}
\end{center}
\end{table}
\end{small}

\begin{figure}[t]
	\begin{center}
		\includegraphics[width=1\linewidth]{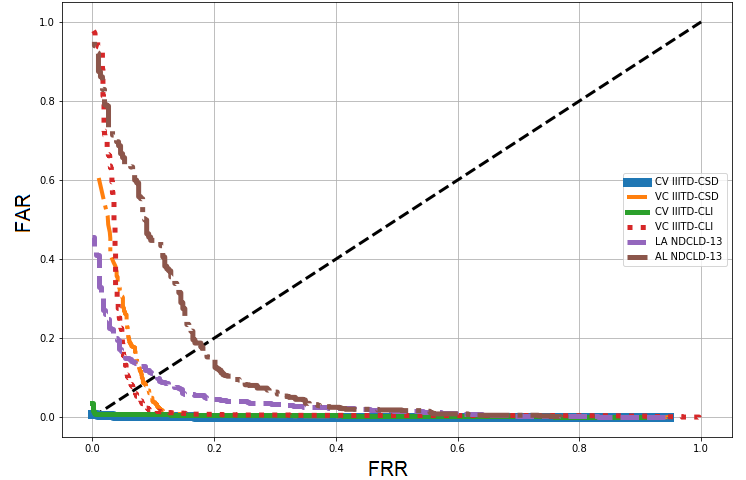}
	\end{center}
	\caption{DET plot for comparing different inter-sensor experiments}
	\label{fig:det2}
\end{figure}

\textbf{2) Inter-Sensor Experiments:}
Inter-sensor experiments have been conducted to validate the performance of the proposed model in environments contrastive to the training.  Training-testing strategy in this experiment involved utilization of entire data from one-sensor in training while from the other is testing. Table \ref{tab:3} tabulates performances achieved by various models under this strategy. It can be concluded from the results that DFCANet is robust enough to capture the intricacies in cross-sensor setting. Specifically, for the case of IIITD-CLI and IIITD-CSD the performance obtained is encouraging. The models tested on Vista-sensor obtain ACER as low as 0.56\% and 0.67\% respectively. Figure \ref{fig:det2} illustrates the  comparative performance of inter-sensor based experimental results using DET curves. Similarly, the average accuracies attained by the same is significantly higher. In compared to the Intra-Sensor the performance drops in challenging experimental setup of are not stark at all. A point to highlight is that, unlike Intra-Sensor experiments the model attains superlative performance in IIITD-CSD database when compared with IIITD-CLI database. It can also be inferred that for most of the models, APCER is relatively higher than the NPCER. Due to limited data quantities for AD100 sensor, though training data is used completely, but while testing the same trained model on LG4000, 900 random examples from the LG400 dataset is sampled. It can be observed from Table, for NDCLD’13 database, cross-scenario is challenging for the model although the average accuracies obtained are significantly high.  

\begin{small}
\begin{table}[ht]\scriptsize
\caption{Testing Performance on combined-sensor experiments in terms of AA (\%) APCER (\%), NPCER (\%), and ACER (\%)}
	\label{tab:3}
\begin{center}
\begin{tabular}{|l|l|l|l|l|}
\hline
Dataset   & AA & APCER & NPCER & ACER \\ \hline
IIITD-CSD & 98.02    & 2.43  & 1.54  & 1.99 \\ \hline
IIITD-CLI & 99.09    & 1.69  & 0.51  & 1.10 \\ \hline
NDCLD13   & 93.00    & 8.00  & 6.50  & 7.25 \\ \hline
\end{tabular}
\end{center}
\end{table}
\end{small}

\textbf{3) Combined-Sensor Experiments:} Combined-Senor experiments involve training the model on 50\% data of available sensors while in testing the remaining 50\% of the data is used. These experiments elucidate the overall performance attained by a model in a specific database. The results from this experiment have been tabulated in Table \ref{tab:4}. Performance of the IIITD-CLI trained model is the most optimal. A low ACER of 1.10\% and high accuracy of 99.09\% for the same represents the above-forth mentioned. For IIITD-CSD, and NDCLD’13, the respective performances obtained are significantly higher. There is an overall increment for NDCLD’13 database in comparison with inter and intra sensor experiments, the reason for the same is expanded and more directed training data. Similar, to previous models APCER remains higher than NPCER, and hence it can be concluded that the model is more robust in classifying attacks.
\begin{small}
\begin{table}[ht]\scriptsize
\caption{Testing Performance on incremental learning experiments in terms of AA (\%) APCER (\%), NPCER (\%), and ACER (\%) }
	\label{tab:4}
\begin{center}
\begin{tabular}{|l|l|l|l|l|l|l|}
\hline
Experiment                    & Training                                          & Testing                                          & AA & APCER  & NPCER & ACER  \\ \hline
Intra-sensor& AD100                                                         & AD100                                                         & 80.33             & 36.00     & 11.50     & 23.75    \\ \cline{2-7} 
                              & LG4000                                                        & LG4000                                                        & 94.66             & 6.75      & 5.62      & 5.68     \\ \hline
Inter-sensor & AD100                                                         & LG4000                                                        & 85.44             & 15.18     & 14.23     & 14.70    \\ \cline{2-7} 
                              & LG4000                                                        & AD100                                                         & 90.22             & 21.60     & 4.16      & 12.53    \\ \hline
Combined                      & \begin{tabular}[c]{@{}l@{}}AD100+\\    LG4000\end{tabular} & \begin{tabular}[c]{@{}l@{}}AD100+\\     LG4000\end{tabular} & 93.93             & 11.20     & 3.50      & 7.35     \\ \hline
\end{tabular}
\end{center}
\end{table}
\end{small}

\textbf{4) Incremental Learning Experiments:}
To bring gains in performance for NDCLD’13, we leveraged IIITD-CLI pretrained model to be fine-tuned with NDCLD’13. As both IIITD-CLI and NDCLD’13 is composed of congruent characteristics, hence the model trained on the most diverse data i.e., the combined-sensor model was chosen for inductive-transfer of its weights. After, the transfer fine-tuning on inter, intra and combined sensor datasets of NDCLD13 was done. Result achieved in the experiment is presented in Table \ref{tab:4}.
From the table, when compared to setting without inductive transfer, a prominent gain of 4.33\% in average accuracy for the Intra-Sensor experiment of AD100 is observed. Similarlily, there is increment for the combined setting. Then again, there is a stark incremental gain in ACER of about 2.27\% for the Inter-Sensor experiment of AD100 training and LG4000 testing. These enhancements in performance can be accredited to transfer of iris-feature understandings by the IIITD-CLI Combined sensor model. Nevertheless, there is a slight decrement in performance for LG4000 model for Intra and Inter sensor experimentation. A possible reason can be difference in morphological and textural properties of data in LG4000 and IIITD-CLI. 

\textbf{5) Cross-database Experiments:} 
To validate efficacy in the proposed model, we have conducted experiments under cross-database setting. Cross-Database experiments are a practical-measure of generalizability of the model in the ‘wild’ setting. In this experiment set, use of NDCLD’13 and IIITD-CLI database has been done and the reason being similar characteristics for attacks between the two. The experimental-setup involved training the model with complete data of three sensors while testing on remaining one. Table \ref{tab:6}, encompasses the results obtained in this experiment. The highest performance in this setting is attained when Vista sensor is evaluated while the other three are tested. The following model attains 98.66\% correct classifications while maintaining a low ACER of 1.21\%. For the model tested on Cogent sensor, a decrement in ACER of about 1.16\% is observed in comparison with Inter-Sensor training-testing of Vista-Cogent Sensor. It can be concluded from the table that whenever testing is performed on IIITD-CLI database results are significant but for testing on any of the sensors from NDCLD’13 the performance remains challenging. From the results obtained in Incremental Learning Experiments and Cross-Database experiments, it can be inferred that LG4000 bears different image properties than the other three. This is not the case with AD100 sensor, as including it in training helps the model overcoming training-subtleties, in-turn leading to enhanced performance. Furthermore, it can also be noticed from the table that APCER is significantly higher than NPCER for models wherein testing is conducted on sensors from NDCLD-13. 

\begin{small}
\begin{table}[ht]\scriptsize
\caption{Testing Performance on cross-dataset experiments in terms of AA (\%) APCER (\%), NPCER (\%), and ACER (\%)}
	\label{tab:6}
\begin{center}
\begin{tabular}{|l|l|l|l|l|l|}
\hline
Training                                                             & Testing & AA & APCER & NPCER & ACER  \\ \hline
\begin{tabular}[c]{@{}l@{}}Cogent+Vista+\\   LG4000\end{tabular} & AD100   & 84..44   & 37.33 & 4.66  & 21.00 \\ \hline
\begin{tabular}[c]{@{}l@{}}Cogent+Vista+\\  AD100\end{tabular}  & LG4000  & 73.76    & 67.68 & 5.46  & 36.22 \\ \hline
\begin{tabular}[c]{@{}l@{}}Cogent+AD100+\\   LG4000\end{tabular} & Vista   & 98.66    & 0.94  & 1.48  & 1.21  \\ \hline
\begin{tabular}[c]{@{}l@{}}Vista+LG4000+\\   AD100\end{tabular}  & Cogent  & 95.71    & 6.21  & 3.20  & 4.71  \\ \hline
\end{tabular}
\end{center}
\end{table}
\end{small}

\subsection{Ablation Study on use of FCNet and CAM}
In order to validate contribution in performance by each of the components of DFCANet, an ablation study has been conducted. We removed one component at a time and trained the model over the challenging NDCLD’13 dataset. Results of this study can be referenced from Table \ref{tab:7v}.
It can be inferred from the following table that DFCANet outperforms all other models. It is optimal in terms of average accuracy by 0.60\% than the second-best performing model. From these, results significance of FCNet can be observed, whenever it is removed there is a degradation in performance. Also, FCNet further contributes in lowering the ACER.

\begin{small}
\begin{table}[ht]\scriptsize
\caption{Ablation study on NDCLD'13 dataset in terms of AA (\%) APCER (\%), NPCER (\%), and ACER (\%)}
	\label{tab:7v}
	\begin{center}
\begin{tabular}{|l|l|l|l|l|}
\hline
Model            & AA & APCER & NPCER & ACER \\ \hline
DenseNet         & 92.20    & 7.00  & 8.20  & 7.60 \\ \hline
DenseNet + FCNet & 92.40    & 5.20  & 8.80  & 6.99 \\ \hline
DenseNet + CAM   & 90.20    & 10.20 & 9.60  & 9.90 \\ \hline
DFCANet          & 93.00    & 8.00  & 6.50  & 7.25 \\ \hline
\end{tabular}
\end{center}
\end{table}
\end{small}

\subsection{Feature Map Visualization and analysis}
\begin{figure}[t]
	\begin{center}
		\includegraphics[width=1\linewidth]{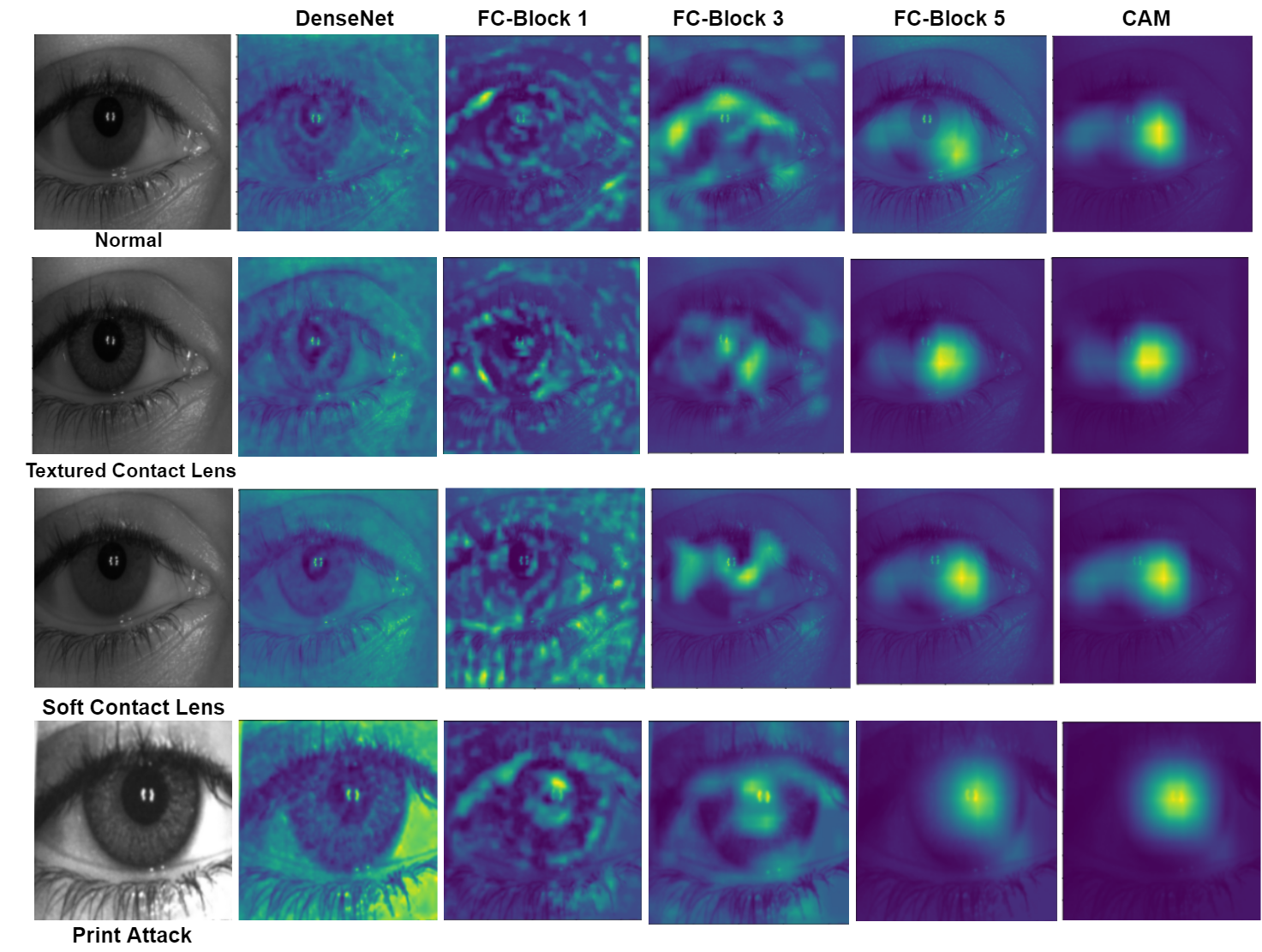}
	\end{center}
	\caption{Iris feature map visualizations for bonafide and attack samples in the IIITD-CLD testing set. The column from left to right refers to the raw samples, maps produced by DenseNet, FCNet-1, FCNet-3, FCNet-5, and CAM  respectively. The first row is the bonafide samples and the second rows are textured contact lens, third rows are soft contact lens, and last row depicts print attacks}
	\label{fig:5a}
\end{figure}

In order to show the learning details of DFCANet, we arbitrary choose iris test images of IIITD-CLD dataset. Figure \ref{fig:5a} visually analyzes the nuances at differnt abstractions of proposed model. To serve the purpose, we have extracted the output feature maps of normal as well its corresponding images from various attacks. Then, the resized feature maps have been plotted with their respective original images. This gives insights on model’s attention at each of the layer. From Figure \ref{fig:5a}, it has been observed that, DenseNet extract generalized features at global scale. Then, within consequent layers of the FCNet, domain-specific feature learning is performed. In the first block of FC-Net, output is activated both locally and globally. As the FCNet gets deeper, the output becomes more prominent over the dominant iris-patterns. Finally, the output of FCNet is generated with high-attention over the patterns that are generated from local-global interactions. These features are further enhanced using the CAM. Furthermore, from the maps it is clear that DFCANet facilitates attention onto iris-region and hence being robust against the occlusions and morphological hindrances present in the ocular images. From the visualizations it can also be concluded that the patterns in the iris-images are the key distinguishing factors between bonafide and adversarial attempts but DFCANet is dedicated in learning representations specifically from the iris-region. From comparing the maps and raw images of normal and soft-lens, similarity in intrinsic learning and attention of the model along with correlated patterns between the two can be inferred. This is a visual-qualitative proof over similar visual-semantics present within the two.

\begin{small}
\begin{table}[ht]\scriptsize
\caption{Intra-sensor comparison for IIITD-CLI dataset in terms of AA (in \%)}
	\label{tab:7x}
	\begin{center}
\begin{tabular}{|l|l|l|l|l|l|}
\hline
Sensor & \begin{tabular}[c]{@{}l@{}}LBP\\ +SVM \cite{gupta2014iris} \end{tabular} & mLBP \cite{yadav2014unraveling} & MVANet \cite{gupta2021generalized} & \begin{tabular}[c]{@{}l@{}}DCCNet\\ \cite{choudhary2020iris} \end{tabular} & \begin{tabular}[c]{@{}l@{}}Proposed\\ DFCANet\end{tabular} \\ \hline
Cogent & 77.46                                                            & 80.87        & 94.90          & 98.71                                                                   & 98.90                                                      \\ \hline
Vista  & 76.01                                                            & 93.97        & 95.91          & 99.30                                                                   & 99.79                                                      \\ \hline
\end{tabular}
\end{center}
\end{table}
\end{small}

\subsection{t-SNE Representation and Analysis}
\begin{figure*}[!hbtp]
\small
\centering
\subfloat[Cogent-Cogent IIITD-CSD ]{\includegraphics[scale = 0.20]{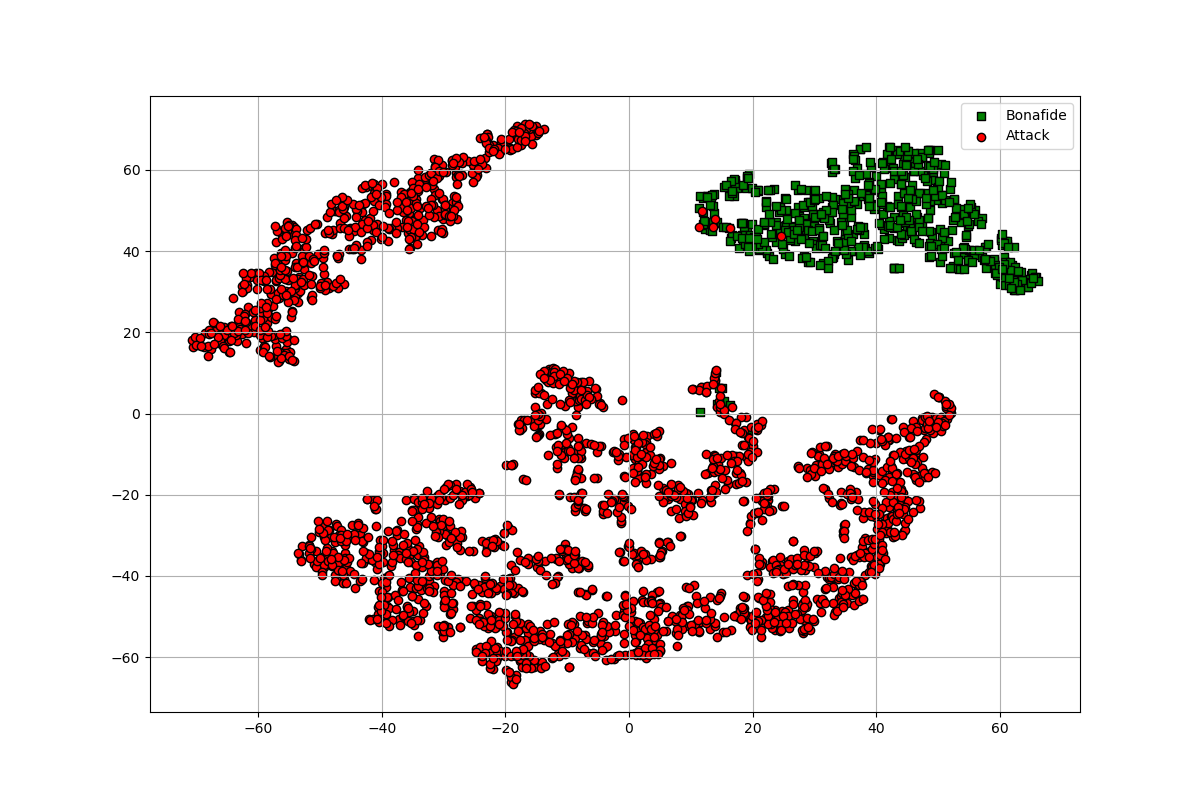}} %
\subfloat[Vista-Vista IIITD-CLI ]{\includegraphics[scale = 0.20]{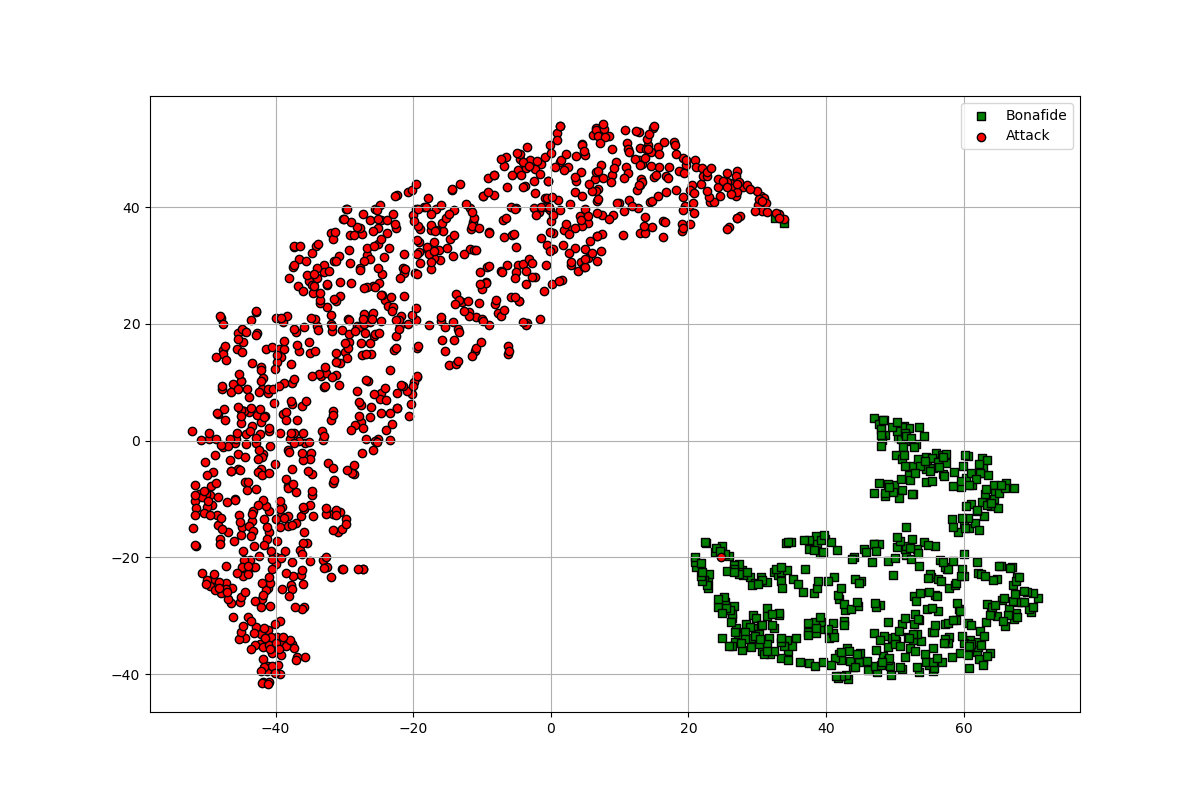}} %
\subfloat[LG4000-LG4000 NDCLD’13 ]{\includegraphics[scale = 0.20]{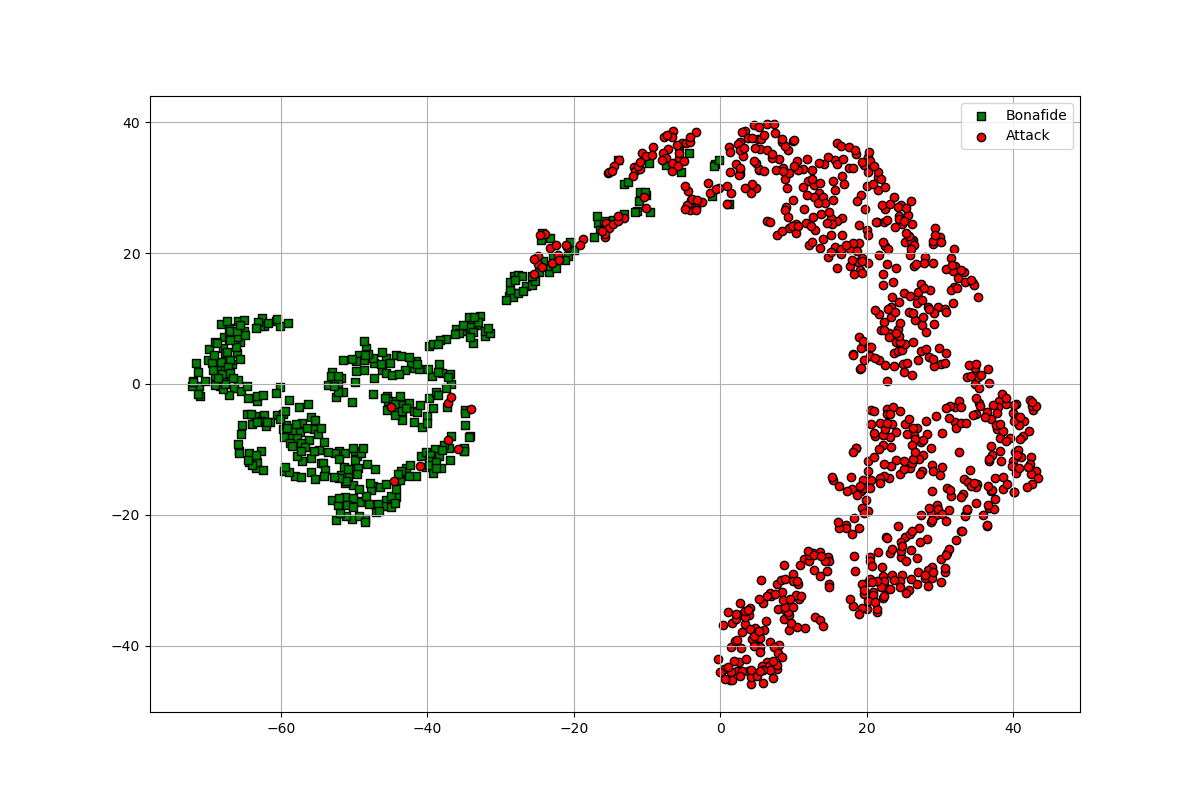}} %
\\
\subfloat[Cogent-Vista IIITD-CSD ]{\includegraphics[scale = 0.19]{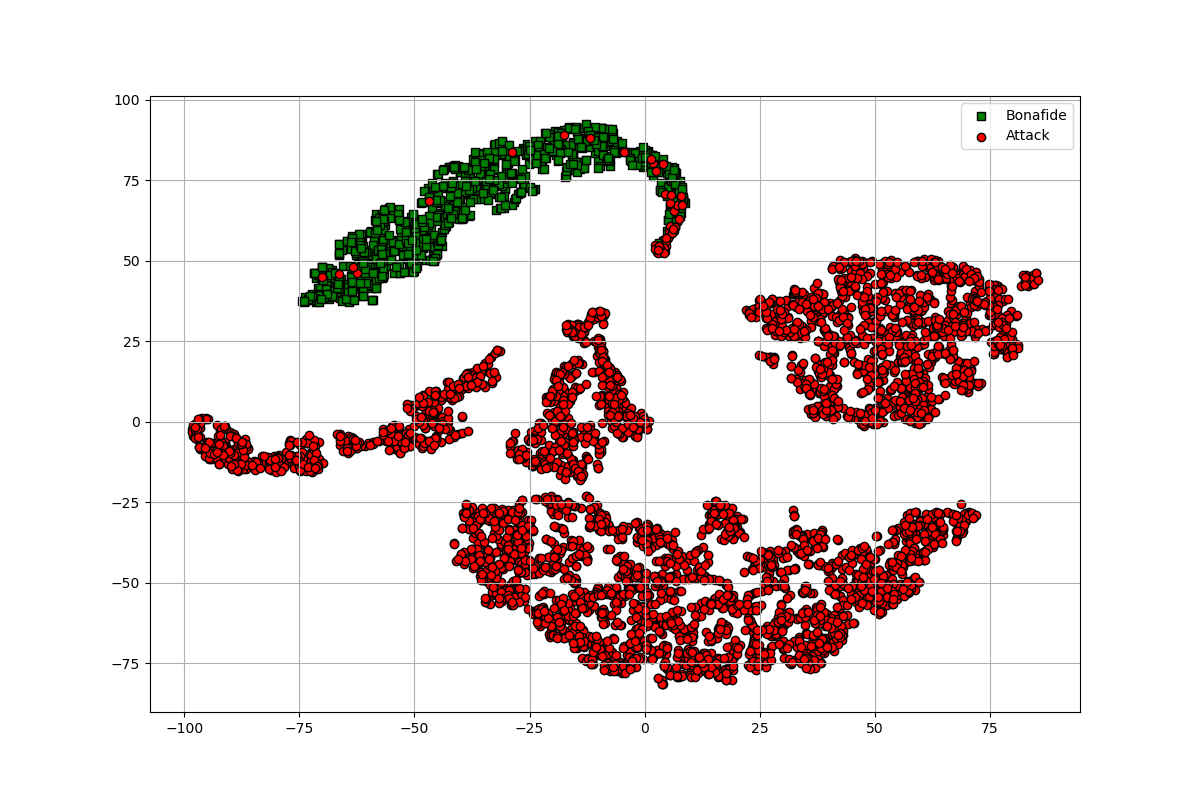}}%
\hspace{0.01 em}
\subfloat[Vista-Cogent IIITD-CLI]{\includegraphics[scale = 0.19]{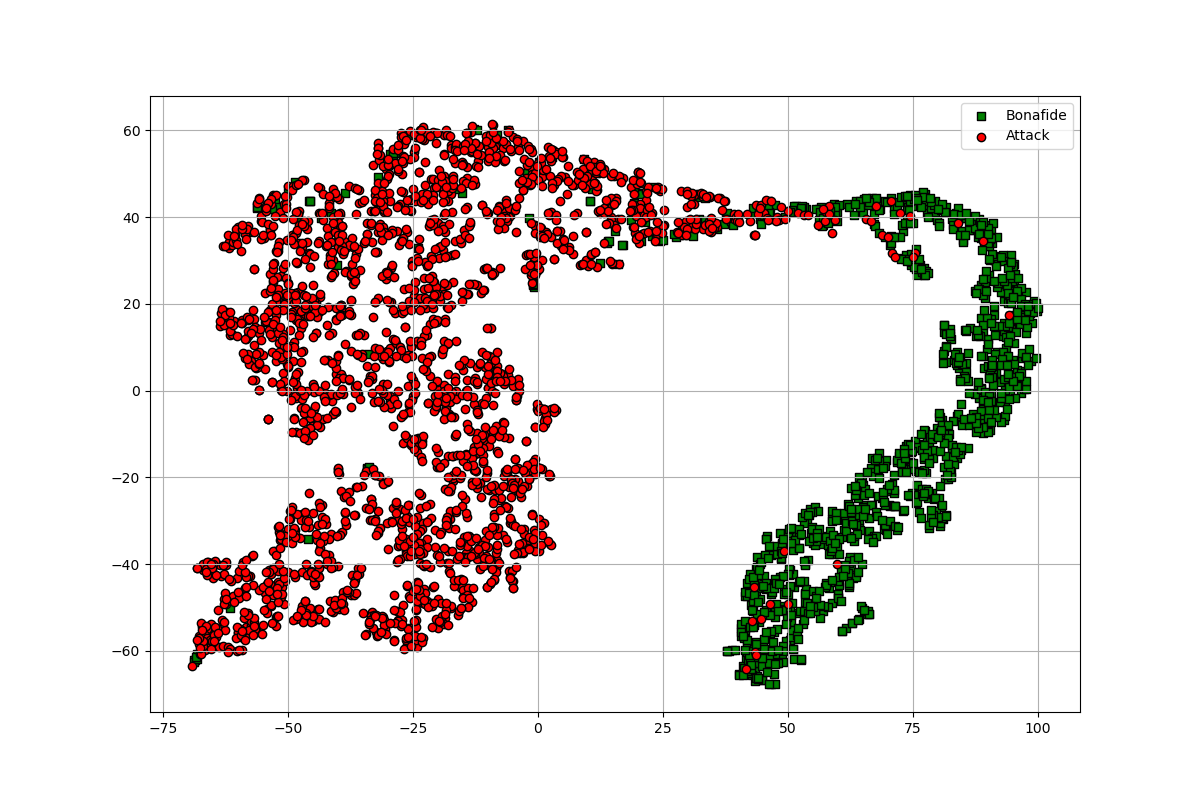}}%
\hspace{0.01 em}
\subfloat[LG4000-AD100 NDCLD’13]{\includegraphics[scale = 0.19]{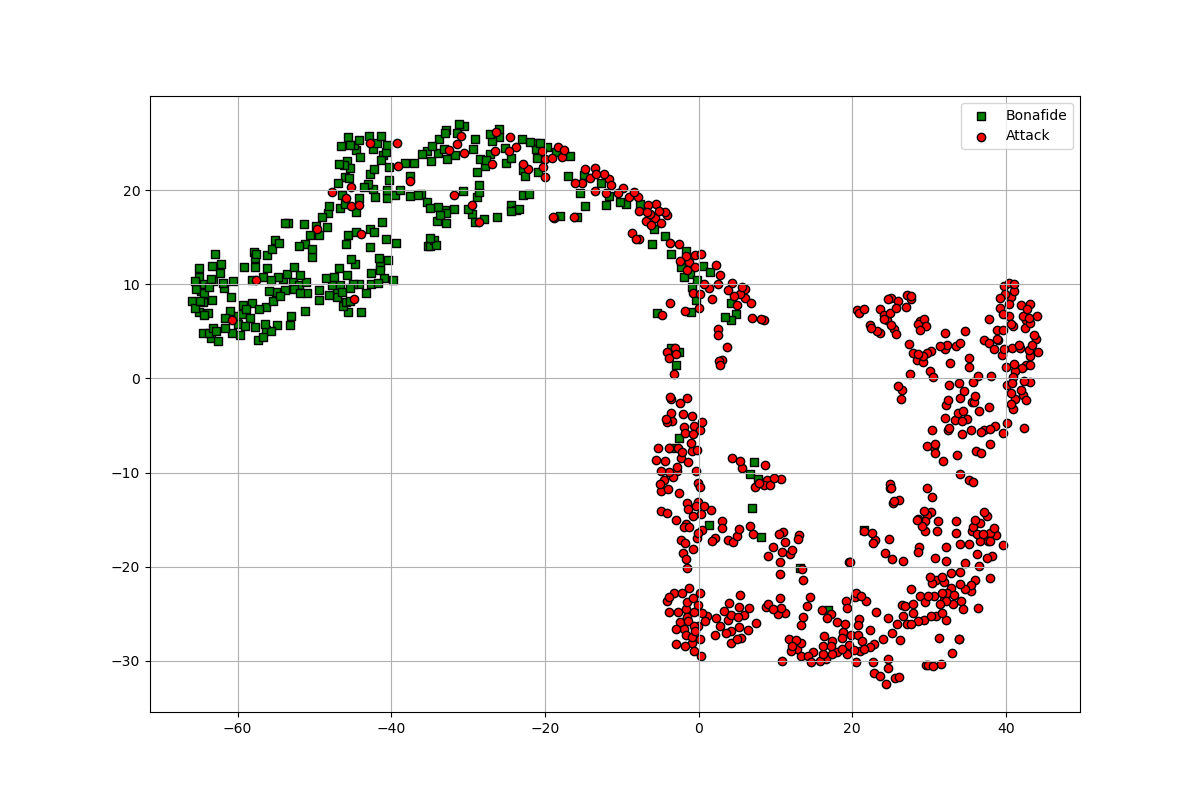}}%
\caption{Illustration of data modeling abilities of proposed DFCANet for Intra sensor (a, b, c) and Inter sensor (d, e, f) experiments conducted over all the three datasets: IITD-CSD, IIITD-CLI and NDCLD'13}
\label{fig:tsne}
\end{figure*}

A visual demonstration over decision-boundaries and class-discrepancies has been presented in Figure \ref{fig:tsne} using t-SNE plots corresponding to models of Intra and Inter sensor experiments. From the trained models, embeddings of the test-set were extracted from the last fully connected layer and were fed to t-SNE algorithm. From the illustrated t-SNE plots, stark separative boundaries between bonafide and attack samples can be observed. 
Both attack and bonafide embeddings are tightly
clustered within themselves while variance gets robustly modelled. Also, multi-center nature in the attack samples can be inferred in the plots. The following multimodal characteristics within attack is induced due to presence of soft-textured lenses in the class. This implies to the challenge that is added due to the above-forth mentioned inclusion but at the
same time the well separated plots give evidences over the robustness of the proposed model under challenging protocol. As shown, the t-SNE plots in Figure \ref{fig:tsne} (a, b, c) of intra-sensor experiments 
have more distinctive nature in comparison to inter-sensor plots in in Figure \ref{fig:tsne} (d, e, f). The reason for the same is difference image-capture dynamics between the training and testing dataset for the case of inter-sensor experiment. With respect to the datasets, NDCLD’13 appears slightly intricate for the model to capture non-linearities in it while it is quite evident from the same plots that the model is well generalized for cross-sensor evaluation.

\subsection{Comparative analysis}
This section presents comparison of the proposed DFCANet
with the state-of-the-art frameworks. Comparison is conducted
individually for all the three Datasets considered in this study. As mentioned earlier, unlike the standard approaches \cite{yadav2018fusion, fang2021iris} placing soft-contact lens as a bonafide attempt, we considered it a spoofing attack. Since both normal and its corresponding soft-lens based spoof bears similar visual appearance a challenging is posed for the network. The purpose of the same is to maintain integrity and generalizability of the proposed IPAD against any form of attack.
Table \ref{tab:7x} reports the comparative-results for intra
sensor conducted on IIITD-CLI database. Average accuracy
over Intra-Sensor experiment of the PAD system has been
considered as the comparative-metric by most of the previous
works, hence we have considered the same. It
can be clearly seen that DFCANet outperforms the previous approaches by a significant gap. There is gain in performance of about 0.5\%
 in Vista when compared with the state-of-the-art framework.
Further, when compared with the optimal machine learning
based method, a stark gain of 5.82\%, while for the case of
Cogent sensor this quantity rises to 18.03\%. It is essential to mention that the proposed DFCANet has been performing at par with current state-of-the-art even under the hard-testing
protocol of considering soft-lenses as attacks. The high-end
performance achieved by the DFCANet can be accredited to its
capacity to learn locally inherent iris textures and structure with respect to globally spread patterns. The following comparative study further reinforces the fact that Cogent sensor’s captured images are slightly more challenging to classify when compared with corresponding Vista counterpart.

\begin{small}
\begin{table}[ht]\scriptsize
\caption{Combined-Sensor comparison for IIITD-CSD Database in term of ACER (\%)}
	\label{tab:7y}
	\begin{center}

\begin{tabular}{|l|l|l|l|l|}
\hline
Method & wLBP\cite{zhang2010contact}  & DESIST\cite{kohli2016detecting} & MVHF\cite{yadav2018fusion} & \begin{tabular}[c]{@{}l@{}}DFCANet\\ (proposed)\end{tabular} \\ \hline
ACER   & 19.04 & 12.49  & 1.30 & 1.99                                                         \\ \hline
\end{tabular}
\end{center}
\end{table}
\end{small}

For comparing the obtained results with previous works \cite{zhang2010contact, kohli2016detecting, yadav2018fusion} on the challenging database IIIT-CSD, the comparison over combined sensor have been conducted keeping ACER as the choice of performance metric. As depicted in Table \ref{tab:7y}, the average classification rate for the proposed model is as low as 1.99\%. From this result, it can be inferred that even under this challenging strategy, the proposed model performs comparatively with the state-of-the-art methods. Further, there is a 10.50\% decrement in the error rates of the proposed model when it is compared with machine learning models.

\begin{small}
\begin{table*}[ht]\scriptsize
\footnotesize
\caption{Intra-Sensor Comparative IPAD results on
NDCLD’13 }
	\label{tab:7p}
	\begin{center}
\begin{tabular}{|l|l|l|l|l|l|l|l|l|}
\hline
Sensor                  & \begin{tabular}[c]{@{}l@{}}Metric\\ (\%)\end{tabular} & \begin{tabular}[c]{@{}l@{}}wLBP\\ \cite{zhang2010contact}\end{tabular}  & \begin{tabular}[c]{@{}l@{}}DESIST\\ \cite{kohli2016detecting}\end{tabular} & \begin{tabular}[c]{@{}l@{}}MVHF\\ \cite{yadav2018fusion}\end{tabular}& \begin{tabular}[c]{@{}l@{}}A-PBS\\ \cite{fang2021iris}\end{tabular}& \begin{tabular}[c]{@{}l@{}}MSA\\ \cite{fang2021cross}\end{tabular} & \begin{tabular}[c]{@{}l@{}}Proposed DFCANet\\ (Soft Lens as Bonafide)\end{tabular} & \begin{tabular}[c]{@{}l@{}}Proposed DFCANet\\ Soft Lens as Attack\end{tabular} \\ \hline
LG4000 & APCER                                                & 2.00  & 0.50   & 0.00 & 0.00 & 0.00  & 0.00                                                                               & 9.75                                                                           \\ \cline{2-9} 
                        & NPCER                                                & 1.00  & 0.50   & 0.00 & 0.00 & 0.00  & 0.25                                                                               & 1.25                                                                           \\ \cline{2-9} 
                        & ACER                                                 & 1.50  & 0.50   & 0.00 & 0.00 & 0.00  & 0.12                                                                               & 5.50                                                                           \\ \hline
AD100  & APCER                                                & 9.00  & 2.00   & 1.00 & 0.00 & 1.00  & 0.00                                                                               & 44.00                                                                          \\ \cline{2-9} 
                        & NPCER                                                & 14.00 & 1.50   & 0.00 & 0.00 & 0.00  & 0.00                                                                               & 14.00                                                                          \\ \cline{2-9} 
                        & ACER                                                 & 11.50 & 1.75   & 0.50 & 0.00 & 0.50  & 0.00                                                                               & 29.00                                                                          \\ \hline
\end{tabular}
\end{center}
\end{table*}
\end{small}

In NDCLD’13 the performance of the proposed DFCANet has not been significantly high in comparison with the previous frameworks \cite{yadav2018fusion, fang2021iris} that have been considering soft-contact lens as a bonafide attempt. Hence, we have first compared our results with state-of-the-art IPAD approaches \cite{yadav2018fusion, fang2021iris, fang2021cross} including soft-contact lens under attack and bonafide
attempts respectively. Secondly to further verify
generalizability of our model, we considered contact lens detection task in account and compared the proposed model with existing works in the literature \cite{raghavendra2017contlensnet, singh2018ghclnet}. 
For this part of study, we have considered NDCLD’13 database for training and testing. Data from both LG4000 and AD100 was combined and split into two
equal-halves for three class classification.

In Table \ref{tab:7p}, results of PAD on NDCLD’13 database have been compared with existing works. Comparison has been done for intra-sensor experiment using APCER, NPCER and ACER. It is evident from the results that the model considering soft-contact lens as attack attempt finds it utmost challenging to differentiate between a soft-lens and corresponding normal images. This fact is further highlighted numerically with stark 5.38\% and 29.00\% decrement in ACER for LG4000 and AD100 sensor respectively, when compared with the proposed model considering soft-lenses as bonafide attempt. Further, it is also clear that under the following challenging protocol, the proposed DFCANet is generalizing well for the LG4000 sensor but due to limited training data quantities for the AD100 sensor performance gets degraded. Nevertheless, when soft-lenses are considered as bonafide attempt, the proposed DFCANet attains state-of-the-art results. For AD100 sensor there is 0.00\% error while for LG4000 error remains as low as 0.12\%. Since, the model is designed in accordance with task taking soft-lens, the performance of the proposed model for PAD with soft-lenses as bonafide is sub-optimal and can be enhanced with dedicated hyperparametrization.

\begin{figure}[t]
	\begin{center}
		\includegraphics[width=1\linewidth, height=7cm]{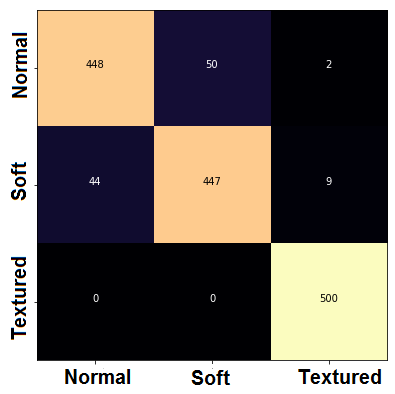}
	\end{center}
	\caption{Confusion matrix plot of DFCANet on contact-lens detection}
	\label{fig:b}
\end{figure}

In Table \ref{tab:7z}, we have tabulated the results of the comparison for contact-lens detection under combined-sensor experiments. Since, all the previous attempts for contact-lens detection has been using class-based matching, \cite{choudhary2020iris} involved validation using three class-classification, thereby we have reported result in it. From the Table 12, it is evident that DFCANet outperforms previous methods by 3-8\% in performance. To get further insights on where the model is facing challenges, we plotted confusion matrix of DFCANet trained for contact-lens detection, same can referenced from Figure \ref{fig:b}.
One can understand that our model is performing significantly well in classifying textured contact lens (i.e., 100\% detection rate) but is having weaker decision boundary between soft contact lens and normal. This further implies that there is stringent enforcement on model to differentiate between normal and soft contact lens, and it is apparently facing challenges in the same. Nevertheless, the high-end results in overall comparison suggests model’s invariant robustness against various types of presentation attacks.

\begin{small}
\begin{table}[ht]\scriptsize
\caption{Combined-Sensor Contact Lens Detection results on NDCLD'13}
	\label{tab:7z}
	\begin{center}

\begin{tabular}{|l|l|l|l|l|}
\hline
Method    & ContlensNet\cite{raghavendra2017contlensnet} & GHCLNet\cite{singh2018ghclnet} & DCCNet\cite{choudhary2020iris} & \begin{tabular}[c]{@{}l@{}}DFCANet\\ (proposed)\end{tabular} \\ \hline
Avg. Acc. & 85.45       & 87.01   & 90.04  & 93.00                                                        \\ \hline
\end{tabular}

\end{center}
\end{table}
\end{small}

\section{Conclusions}
Existing IPAD algorithms are burdened with penalties on performance under unseen and cross-domain scenarios which do not make them ideal for newer formats of iris-based presentation attacks. This paper presented DFCANet, a novel domain-specific deep learning model that obtained state-of-the-art results under various strenuous conditions. Experiments conducted on the IIITD-CSD, IIITD-CLI, and NDCLD'13 datasets show that the proposed network is not only effective in intra-domain settings but generalizes well enough for various cross-domain scenarios. Further, we observed drastic performance degradations when soft-lenses are considered under attack category. Under such challenging protocols, a significant gain of 0.5\% is observed for IIITD-CLI Vista while for IIITD-CLI, average error rate as low as 1.99\% is obtained.  Under standard IPAD evaluation strategy on NDCLD’13, the proposed model achieves 0\% and 0.12\% error rates for AD100 and LG4000 sensors respectively. To address further shortcomings, a novel incremental learning based methodology was introduced which brought significant 4.33\% gain in terms of average accuracy. In future work, more challenging and adverse evaluation scenarios will be considered that will involve unseen and new-forms of attacks. Evaluation of IPAD algorithms on cross-user and open-set settings shall be another major aspect for future research. 

\bibliographystyle{IEEEtran}
\bibliography{ref}{}

\begin{thebibliography}{10}
\providecommand{\url}[1]{#1}
\csname url@samestyle\endcsname
\providecommand{\newblock}{\relax}
\providecommand{\bibinfo}[2]{#2}
\providecommand{\BIBentrySTDinterwordspacing}{\spaceskip=0pt\relax}
\providecommand{\BIBentryALTinterwordstretchfactor}{4}
\providecommand{\BIBentryALTinterwordspacing}{\spaceskip=\fontdimen2\font plus
\BIBentryALTinterwordstretchfactor\fontdimen3\font minus
  \fontdimen4\font\relax}
\providecommand{\BIBforeignlanguage}[2]{{%
\expandafter\ifx\csname l@#1\endcsname\relax
\typeout{** WARNING: IEEEtran.bst: No hyphenation pattern has been}%
\typeout{** loaded for the language `#1'. Using the pattern for}%
\typeout{** the default language instead.}%
\else
\language=\csname l@#1\endcsname
\fi
#2}}
\providecommand{\BIBdecl}{\relax}
\BIBdecl

\bibitem{jain201650}
A.~K. Jain, K.~Nandakumar, and A.~Ross, ``50 years of biometric research:
  Accomplishments, challenges, and opportunities,'' \emph{Pattern recognition
  letters}, vol.~79, pp. 80--105, 2016.

\bibitem{garg2019toward}
S.~Garg, K.~Kaur, G.~Kaddoum, and K.-K.~R. Choo, ``Toward secure and provable
  authentication for internet of things: Realizing industry 4.0,'' \emph{IEEE
  Internet of Things Journal}, vol.~7, no.~5, pp. 4598--4606, 2019.

\bibitem{prabhakar2003biometric}
S.~Prabhakar, S.~Pankanti, and A.~K. Jain, ``Biometric recognition: Security
  and privacy concerns,'' \emph{IEEE security \& privacy}, vol.~1, no.~2, pp.
  33--42, 2003.

\bibitem{jaswal2016knuckle}
G.~Jaswal, A.~Kaul, and R.~Nath, ``Knuckle print biometrics and fusion
  schemes--overview, challenges, and solutions,'' \emph{ACM Computing Surveys
  (CSUR)}, vol.~49, no.~2, pp. 1--46, 2016.

\bibitem{jain2021biometrics}
A.~K. Jain, D.~Deb, and J.~J. Engelsma, ``Biometrics: Trust, but verify,''
  \emph{arXiv preprint arXiv:2105.06625}, 2021.

\bibitem{drozdowski2020demographic}
P.~Drozdowski, C.~Rathgeb, A.~Dantcheva, N.~Damer, and C.~Busch, ``Demographic
  bias in biometrics: A survey on an emerging challenge,'' \emph{IEEE
  Transactions on Technology and Society}, vol.~1, no.~2, pp. 89--103, 2020.

\bibitem{uludag2004attacks}
U.~Uludag and A.~K. Jain, ``Attacks on biometric systems: a case study in
  fingerprints,'' in \emph{Security, steganography, and watermarking of
  multimedia contents VI}, vol. 5306.\hskip 1em plus 0.5em minus 0.4em\relax
  International Society for Optics and Photonics, 2004, pp. 622--633.

\bibitem{tolosana2019biometric}
R.~Tolosana, M.~Gomez-Barrero, C.~Busch, and J.~Ortega-Garcia, ``Biometric
  presentation attack detection: Beyond the visible spectrum,'' \emph{IEEE
  Transactions on Information Forensics and Security}, vol.~15, pp. 1261--1275,
  2019.

\bibitem{daugman2003importance}
J.~Daugman, ``The importance of being random: statistical principles of iris
  recognition,'' \emph{Pattern recognition}, vol.~36, no.~2, pp. 279--291,
  2003.

\bibitem{daugman2007new}
------, ``New methods in iris recognition,'' \emph{IEEE Transactions on
  Systems, Man, and Cybernetics, Part B (Cybernetics)}, vol.~37, no.~5, pp.
  1167--1175, 2007.

\bibitem{morales2019introduction}
A.~Morales, J.~Fierrez, J.~Galbally, and M.~Gomez-Barrero, ``Introduction to
  iris presentation attack detection,'' in \emph{Handbook of Biometric
  Anti-Spoofing}.\hskip 1em plus 0.5em minus 0.4em\relax Springer, 2019, pp.
  135--150.

\bibitem{marcel2019handbook}
S.~Marcel, M.~S. Nixon, J.~Fierrez, and N.~Evans, \emph{Handbook of biometric
  anti-spoofing: Presentation attack detection}.\hskip 1em plus 0.5em minus
  0.4em\relax Springer, 2019.

\bibitem{fang2021cross}
M.~Fang, N.~Damer, F.~Boutros, F.~Kirchbuchner, and A.~Kuijper,
  ``Cross-database and cross-attack iris presentation attack detection using
  micro stripes analyses,'' \emph{Image and Vision Computing}, vol. 105, p.
  104057, 2021.

\bibitem{das2020iris}
P.~Das, J.~Mcfiratht, Z.~Fang, A.~Boyd, G.~Jang, A.~Mohammadi, S.~Purnapatra,
  D.~Yambay, S.~Marcel, M.~Trokielewicz \emph{et~al.}, ``Iris liveness
  detection competition (livdet-iris)-the 2020 edition,'' in \emph{2020 IEEE
  International Joint Conference on Biometrics (IJCB)}.\hskip 1em plus 0.5em
  minus 0.4em\relax IEEE, 2020, pp. 1--9.

\bibitem{yambay2017livdet}
D.~Yambay, B.~Becker, N.~Kohli, D.~Yadav, A.~Czajka, K.~W. Bowyer,
  S.~Schuckers, R.~Singh, M.~Vatsa, A.~Noore \emph{et~al.}, ``Livdet iris
  2017—iris liveness detection competition 2017,'' in \emph{2017 IEEE
  International Joint Conference on Biometrics (IJCB)}.\hskip 1em plus 0.5em
  minus 0.4em\relax IEEE, 2017, pp. 733--741.

\bibitem{gupta2014iris}
P.~Gupta, S.~Behera, M.~Vatsa, and R.~Singh, ``On iris spoofing using print
  attack,'' in \emph{2014 22nd international conference on pattern
  recognition}.\hskip 1em plus 0.5em minus 0.4em\relax IEEE, 2014, pp.
  1681--1686.

\bibitem{yadav2018fusion}
D.~Yadav, N.~Kohli, A.~Agarwal, M.~Vatsa, R.~Singh, and A.~Noore, ``Fusion of
  handcrafted and deep learning features for large-scale multiple iris
  presentation attack detection,'' in \emph{Proceedings of the IEEE Conference
  on Computer Vision and Pattern Recognition Workshops}, 2018, pp. 572--579.

\bibitem{fang2021iris}
M.~Fang, N.~Damer, F.~Boutros, F.~Kirchbuchner, and A.~Kuijper, ``Iris
  presentation attack detection by attention-based and deep pixel-wise binary
  supervision network,'' in \emph{2021 IEEE International Joint Conference on
  Biometrics (IJCB)}.\hskip 1em plus 0.5em minus 0.4em\relax IEEE, 2021, pp.
  1--8.

\bibitem{sharma2020d}
R.~Sharma and A.~Ross, ``D-netpad: An explainable and interpretable iris
  presentation attack detector,'' in \emph{2020 IEEE International Joint
  Conference on Biometrics (IJCB)}.\hskip 1em plus 0.5em minus 0.4em\relax
  IEEE, 2020, pp. 1--10.

\bibitem{kohli2016detecting}
N.~Kohli, D.~Yadav, M.~Vatsa, R.~Singh, and A.~Noore, ``Detecting medley of
  iris spoofing attacks using desist,'' in \emph{2016 IEEE 8th International
  Conference on Biometrics Theory, Applications and Systems (BTAS)}.\hskip 1em
  plus 0.5em minus 0.4em\relax IEEE, 2016, pp. 1--6.

\bibitem{kuehlkamp2018ensemble}
A.~Kuehlkamp, A.~Pinto, A.~Rocha, K.~W. Bowyer, and A.~Czajka, ``Ensemble of
  multi-view learning classifiers for cross-domain iris presentation attack
  detection,'' \emph{IEEE Transactions on Information Forensics and Security},
  vol.~14, no.~6, pp. 1419--1431, 2018.

\bibitem{daugman2000wavelet}
J.~Daugman, ``Wavelet demodulation codes, statistical independence, and pattern
  recognition,'' \emph{Institute of Mathematics and its Applications, Proc. 2nd
  IMA-IP: Mathematical Methods, Algorithms, and Applications (Blackledge and
  Turner, Eds)}, pp. 244--260, 2000.

\bibitem{sun2013iris}
Z.~Sun, H.~Zhang, T.~Tan, and J.~Wang, ``Iris image classification based on
  hierarchical visual codebook,'' \emph{IEEE Transactions on pattern analysis
  and machine intelligence}, vol.~36, no.~6, pp. 1120--1133, 2013.

\bibitem{he2009efficient}
Z.~He, Z.~Sun, T.~Tan, and Z.~Wei, ``Efficient iris spoof detection via boosted
  local binary patterns,'' in \emph{International conference on
  biometrics}.\hskip 1em plus 0.5em minus 0.4em\relax Springer, 2009, pp.
  1080--1090.

\bibitem{zhang2010contact}
H.~Zhang, Z.~Sun, and T.~Tan, ``Contact lens detection based on weighted lbp,''
  in \emph{2010 20th International Conference on Pattern Recognition}.\hskip
  1em plus 0.5em minus 0.4em\relax IEEE, 2010, pp. 4279--4282.

\bibitem{hu2016iris}
Y.~Hu, K.~Sirlantzis, and G.~Howells, ``Iris liveness detection using regional
  features,'' \emph{Pattern Recognition Letters}, vol.~82, pp. 242--250, 2016.

\bibitem{gragnaniello2016using}
D.~Gragnaniello, G.~Poggi, C.~Sansone, and L.~Verdoliva, ``Using iris and
  sclera for detection and classification of contact lenses,'' \emph{Pattern
  Recognition Letters}, vol.~82, pp. 251--257, 2016.

\bibitem{raja2015video}
K.~B. Raja, R.~Raghavendra, and C.~Busch, ``Video presentation attack detection
  in visible spectrum iris recognition using magnified phase information,''
  \emph{IEEE Transactions on Information Forensics and Security}, vol.~10,
  no.~10, pp. 2048--2056, 2015.

\bibitem{raghavendra2015robust}
R.~Raghavendra and C.~Busch, ``Robust scheme for iris presentation attack
  detection using multiscale binarized statistical image features,'' \emph{IEEE
  Transactions on Information Forensics and Security}, vol.~10, no.~4, pp.
  703--715, 2015.

\bibitem{menotti2015deep}
D.~Menotti, G.~Chiachia, A.~Pinto, W.~R. Schwartz, H.~Pedrini, A.~X. Falcao,
  and A.~Rocha, ``Deep representations for iris, face, and fingerprint spoofing
  detection,'' \emph{IEEE Transactions on Information Forensics and Security},
  vol.~10, no.~4, pp. 864--879, 2015.

\bibitem{chenexplainable}
C.~Chen and A.~Ross, ``An explainable attention-guided iris presentation attack
  detector,'' in \emph{Proceedings of the IEEE/CVF Winter Conference on
  Applications of Computer Vision}, pp. 97--106.

\bibitem{pala2017iris}
F.~Pala and B.~Bhanu, ``Iris liveness detection by relative distance
  comparisons,'' in \emph{Proceedings of the IEEE Conference on Computer Vision
  and Pattern Recognition Workshops}, 2017, pp. 162--169.

\bibitem{fang2020deep}
M.~Fang, N.~Damer, F.~Boutros, F.~Kirchbuchner, and A.~Kuijper, ``Deep learning
  multi-layer fusion for an accurate iris presentation attack detection,'' in
  \emph{2020 IEEE 23rd International Conference on Information Fusion
  (FUSION)}.\hskip 1em plus 0.5em minus 0.4em\relax IEEE, 2020, pp. 1--8.

\bibitem{yadav2019detecting}
D.~Yadav, N.~Kohli, M.~Vatsa, R.~Singh, and A.~Noore, ``Detecting textured
  contact lens in uncontrolled environment using densepad,'' in
  \emph{Proceedings of the IEEE/CVF Conference on Computer Vision and Pattern
  Recognition Workshops}, 2019, pp. 0--0.

\bibitem{gupta2021generalized}
M.~Gupta, V.~Singh, A.~Agarwal, M.~Vatsa, and R.~Singh, ``Generalized iris
  presentation attack detection algorithm under cross-database settings,'' in
  \emph{2020 25th International Conference on Pattern Recognition
  (ICPR)}.\hskip 1em plus 0.5em minus 0.4em\relax IEEE, 2021, pp. 5318--5325.

\bibitem{kohli2017synthetic}
N.~Kohli, D.~Yadav, M.~Vatsa, R.~Singh, and A.~Noore, ``Synthetic iris
  presentation attack using idcgan,'' in \emph{2017 IEEE International Joint
  Conference on Biometrics (IJCB)}.\hskip 1em plus 0.5em minus 0.4em\relax
  IEEE, 2017, pp. 674--680.

\bibitem{yadav2021cit}
S.~Yadav and A.~Ross, ``Cit-gan: Cyclic image translation generative
  adversarial network with application in iris presentation attack detection,''
  in \emph{Proceedings of the IEEE/CVF Winter Conference on Applications of
  Computer Vision}, 2021, pp. 2412--2421.

\bibitem{czajka2018presentation}
A.~Czajka and K.~W. Bowyer, ``Presentation attack detection for iris
  recognition: An assessment of the state-of-the-art,'' \emph{ACM Computing
  Surveys (CSUR)}, vol.~51, no.~4, pp. 1--35, 2018.

\bibitem{yambay2019review}
D.~Yambay, A.~Czajka, K.~Bowyer, M.~Vatsa, R.~Singh, A.~Noore, N.~Kohli,
  D.~Yadav, and S.~Schuckers, ``Review of iris presentation attack detection
  competitions,'' in \emph{Handbook of biometric anti-spoofing}.\hskip 1em plus
  0.5em minus 0.4em\relax Springer, 2019, pp. 169--183.

\bibitem{huang2017densely}
G.~Huang, Z.~Liu, L.~Van Der~Maaten, and K.~Q. Weinberger, ``Densely connected
  convolutional networks,'' in \emph{Proceedings of the IEEE conference on
  computer vision and pattern recognition}, 2017, pp. 4700--4708.

\bibitem{krizhevsky2017imagenet}
A.~Krizhevsky, I.~Sutskever, and G.~E. Hinton, ``Imagenet classification with
  deep convolutional neural networks,'' \emph{Communications of the ACM},
  vol.~60, no.~6, pp. 84--90, 2017.

\bibitem{liu2020improving}
J.-J. Liu, Q.~Hou, M.-M. Cheng, C.~Wang, and J.~Feng, ``Improving convolutional
  networks with self-calibrated convolutions,'' in \emph{Proceedings of the
  IEEE/CVF Conference on Computer Vision and Pattern Recognition}, 2020, pp.
  10\,096--10\,105.

\bibitem{yadav2014unraveling}
D.~Yadav, N.~Kohli, J.~S. Doyle, R.~Singh, M.~Vatsa, and K.~W. Bowyer,
  ``Unraveling the effect of textured contact lenses on iris recognition,''
  \emph{IEEE Transactions on Information Forensics and Security}, vol.~9,
  no.~5, pp. 851--862, 2014.

\bibitem{doyle2013variation}
J.~S. Doyle, K.~W. Bowyer, and P.~J. Flynn, ``Variation in accuracy of textured
  contact lens detection based on sensor and lens pattern,'' in \emph{2013 IEEE
  sixth international conference on biometrics: theory, applications and
  systems (BTAS)}.\hskip 1em plus 0.5em minus 0.4em\relax IEEE, 2013, pp. 1--7.

\bibitem{choudhary2020iris}
M.~Choudhary, V.~Tiwari, and U.~Venkanna, ``Iris anti-spoofing through
  score-level fusion of handcrafted and data-driven features,'' \emph{Applied
  Soft Computing}, vol.~91, p. 106206, 2020.

\bibitem{raghavendra2017contlensnet}
R.~Raghavendra, K.~B. Raja, and C.~Busch, ``Contlensnet: Robust iris contact
  lens detection using deep convolutional neural networks,'' in \emph{2017 IEEE
  Winter Conference on Applications of Computer Vision (WACV)}.\hskip 1em plus
  0.5em minus 0.4em\relax IEEE, 2017, pp. 1160--1167.

\bibitem{singh2018ghclnet}
A.~Singh, V.~Mistry, D.~Yadav, and A.~Nigam, ``Ghclnet: A generalized
  hierarchically tuned contact lens detection network,'' in \emph{2018 IEEE 4th
  International Conference on Identity, Security, and Behavior Analysis
  (ISBA)}.\hskip 1em plus 0.5em minus 0.4em\relax IEEE, 2018, pp. 1--8.

\end{thebibliography}

\vspace{11pt}

\bf{If you will not include a photo:}\vspace{-33pt}
\begin{IEEEbiographynophoto}{John Doe}
Use $\backslash${\tt{begin\{IEEEbiographynophoto\}}} and the author name as the argument followed by the biography text.
\end{IEEEbiographynophoto}

\vfill

\end{document}